\documentclass[a4paper,USenglish]{tgdk-v2021}

\usepackage{tikz}
\usepackage{tikzscale}
\tikzstyle{every node}=[draw, fill=white, shape=circle, inner sep = 1.5pt]
\usepackage{packages/fca}
\usepackage{adjustbox}
\usepackage{mathtools}

\usepackage{booktabs}

\newcommand{\eg}{e.\,g., }
\newcommand{\ie}{i.\,e., }

\newcommand{\fett}[1]{\emph{#1}}  %

\newcommand{\randbemerkung}[1]{} %
\newcommand{\selfcite}[1]{\cite{#1}}
\newcommand{\groupcite}[1]{\cite{#1}}
\newcommand{\NPhard}{$\mathcal{NP}$-hard}
\newcommand{\NPcomplete}{$\mathcal{NP}$-complete}
\newcommand*{\longeq}{:\Longleftrightarrow}

\title{Towards Ordinal Data Science} %

\author{Gerd Stumme\footnote{corresponding author}}{Knowledge \& Data Engineering Group, Research Center for Information System Design \&\\ Department of Electrical Engineering and Computer Science, University of Kassel,
	Germany \and
	\url{https://www.kde.cs.uni-kassel.de/en/stumme}}{stumme@cs.uni-kassel.de}{https://orcid.org/0000-0002-0570-7908}{}%

\author{Dominik Dürrschnabel}{Knowledge \& Data Engineering Group, Research Center for Information System Design \&\\ Department of Electrical Engineering and Computer Science,  University of Kassel,
	Germany \and
	\url{https://www.kde.cs.uni-kassel.de/en/duerrschnabel}}{duerrschnabel@cs.uni-kassel.de}{https://orcid.org/0000-0002-0855-4185}{}%

\author{Tom Hanika}{Institute of Computer Science, University of
	Hildesheim, Germany \and Berlin School of Library and Information
	Science,  Humboldt-Universität zu Berlin, Germany
	\and \url{https://www.ibi.hu-berlin.de/de/institut/personen/hanika}
}{tom.hanika@hu-berlin.de}{https://orcid.org/0000-0002-4918-6374}{}%

\authorrunning{G. Stumme, D. Dürrschnabel and T. Hanika}

\Copyright{Gerd Stumme, Dominik Dürrschnabel, Tom Hanika}



\category{Vision}

\relatedversiondetails{Full Version}{https://arxiv.org/abs/2307.09477} 

\ArticleNo{2}

\Volume{1}
\Issue{1}
\Article{42}
\DateSubmission{Date of submission}
\DateAcceptance{Date of acceptance}
\DatePublished{Date of publishing}
\SectionAreaEditor{TGDK section area editor}

 \nolinenumbers

\begin{document}

\maketitle

\begin{abstract}
  Order is one of the main instruments to measure the relationship between objects in (empirical) data. However, compared to methods that use numerical properties of objects, the amount of ordinal methods developed is rather small. One reason for this is the limited availability of computational resources in the last century that would have been required for ordinal computations. Another reason --- particularly important for this line of research --- is that order-based methods are often seen as too mathematically rigorous for applying them to real-world data. In this paper, we will therefore discuss different means for measuring and ‘calculating’ with ordinal structures --- a specific class of directed graphs --- and show how to infer knowledge from them. Our aim is to establish Ordinal Data Science as a fundamentally new research agenda. Besides cross-fertilization with other cornerstone machine learning and knowledge representation methods, a broad range of disciplines will benefit from this endeavor, including, psychology, sociology, economics, web science, knowledge engineering, scientometrics.
\end{abstract}

\section{Introduction}
Order is a predominant concept for perceiving and organizing our physical and social environment, to infer meaning and explanation from
observation, and to search and rectify decisions. For instance, we admire the highest mountain on earth, observe pecking order among animals, schedule events in time, and structure our organizations, libraries, and diseases hierarchically. The notion of order is deeply embedded in our language, as every adjective gives rise to a comparative (\eg better, more expensive, more beautiful). Specific technical and social processes have been established for dealing with ordinal structures, \eg scheduling routines for aircraft take-offs and production planning, queuing at bus stops, deriving the succession order as depth-first linear extension of the royal family tree, or discussing only the borderline cases in scientific program committees.
These processes, however, are rather task-specific --- there exist only few generic data analysis and machine learning tasks that are particularly tailored for ordinal data in general.

It comes thus as little surprise that \fett{\emph{order} is one of the two main types of relations between objects in empirical data}, together with \emph{proximity/distance} \cite{coombs1964theory}.
While there exists a wide range of scientific work on analysis methods for ordinal data, their amount and coverage is far less than for proximity data. As a consequence, many data analysts resort to methods designed for other data types. We therefore call interested researchers to join us in the aim to establish the new field of \emph{Ordinal Data Science} --- both by organizing existing work in a unifying framework and by extending these approaches: Contributions are sought in two specific areas. Theoretical foundations for analyzing ordinal structures, in particular with respect to knowledge extraction and representation, on the one hand; and algorithmic methods that provide the means to measure and `calculate' with ordinal structures and closely related algebraic notions on the other hand.

A wide range of applied disciplines will benefit from the new field of Ordinal Data Science, a.\,o.\ psychology, web science, knowledge engineering, scientometrics.
Besides these disciplines, ordinal data are related to the large family of relational data which have received high interest of the computer science community in the last years. This is due to developments in fields such as sociology (``relational turn'') \cite{mische2011relational} %
or genetics \cite{goto1997organizing} or epidemiology \cite{Ciavarella2016}, and in particular sociotechnical developments such as the rise of online social networks or knowledge graphs.
This means that, for the analysis of ordinal data, one can benefit from all kinds of measures and methods for relational data, as for instance centrality measures and clustering algorithms for (social) network data, or inductive logic programming.
The specific structure of ordinal data, however, allows additionally to tap on the rich --- but up to date mostly unexploited for data science --- tool-set of mathematical order theory \cite{caspard2012finite} and lattice\footnote{In~Section~\ref{what-are-lattices}, we will recall that lattices are specific order relations with additional structure.} theory~\cite{davey2002introduction}.

We expect Ordinal Data Science to become a scientific field with many ramifications. In this paper, we will discuss --- in an exemplary fashion and to kindle the research field --- five out of many potential subfields as programmatic research tasks:

\medskip\textbf{1) Ordinal Measurement Theory.}
Representational Measurement Theory, the most influential theory of measurement to date, discusses how empirically observed phenomena can be measured. It considers a measurement as a mapping from some empirical relational structure to a numerical relational structure. While there seems to be agreement that the latter could be any mathematical structure that is appropriate for the purpose at hand, almost all theory is centered around the real numbers (and variations) as numerical relational structure. As order is the second main type of relation in empirical data beside proximity/distance, we will discuss, in Section~\ref{sec-ordinal-measurement-theory}, the vision of an order-theoretic version of measurement theory.

\medskip\textbf{2) Ordered Metric Spaces.}
Real-world datasets are typically heterogeneous and contain both ordinal and numerical dimensions.
The latter naturally give rise to a metric. In Section~\ref{sec-metric-ordered-spaces}, we will motivate the need for a theory about the compatibility (or consistency) of an order relation and a metric. To this end, we introduce ordered metric spaces (om-spaces), in the spirit of metric measure spaces. Such a theory should be able to answer a range of foundational questions about the relationship between order relations and metrics, including: a) If a metric has been learned based on information about the order, how consistent is the metric with that order? This will provide a fundamental new quality criterion for order-related machine learning tasks. b) How strong do order and metrics ``agree'' with their view on a dataset? An answer might be highly beneficial to an investigation whether order and metrics reflect the same hidden structure. c) How justified is it to represent an order relation by means of the metric? If, for a given dataset, the answer is satisfactory, this would justify in this case the use of the rich toolkit of numerical data analysis and machine learning methods for analyzing the ordinal structure.

\medskip\textbf{3) Algebraic constructions/decompositions for complexity reduction.}
Algebraic operations provide a large variety of methods for reducing the complexity of ordered sets, in particular lattices. However, they are very sensitive to small perturbations in the data; and approximations and heuristics in the style of data mining are not (yet) considered in algebraic research.
In Section~\ref{sec-algebraic-constructions}, we will discuss how they could be made less sensitive in order to make them applicable to data science tasks.

\medskip\textbf{4) Ordinal Factor Analysis.}
Observed and measured data is often highly correlated and interlinked, being caused by a small amount of factors.   A significant task in data analysis is the identification of these factors.
In Section~\ref{sec-ordinal-factor-analysis}, we  discuss --- in analogy to the classical  factor analysis which is based on linear algebra ---  potential ways for extracting these factors in ordinal data based on the relation product.

\medskip\textbf{5) Visualization, Exploration, and Browsing Ordinal Data.}
The typical means of presenting ordered sets to humans is via line diagrams. Open problems include in particular the specification of HCI-founded, formal optimization criteria for graph drawing (and the development of corresponding layout algorithms), as well as the (semi-)automatic break-down of large ordered sets into smaller, visualizable parts together with suitable means for their interactive exploration. In Section~\ref{sec-exploration}, we will call for the development of  new   paradigms for exploring and browsing ordinal datasets.

\medskip
Before diving in the challenges of these subfields in more detail, we will first discuss the nature of ordinal data and the role of hierarchies in data science in Sections~\ref{sec-general-principle} and \ref{sec-role-of-hierarchies}, resp., and will provide some order-theoretical foundations in Section~\ref{sec-foundations}.

\section{Order as Foundational Principle for Organizing Data}
\label{sec-general-principle}

Quantification with real numbers has been boosted by different factors, including $i$) the development of scientific measuring instruments since the scientific revolution, $ii$) the claim that the social sciences (starting with psychology) should use the same numerical methods which had been successful in natural sciences \cite{ferguson1940quantitative}, and $iii$) nowadays by the instant availability of an enormous range of datasets to almost all aspects of science and everyday life.
\randbemerkung{Quantification of ordinal data with real numbers}
Indeed, in many cases, entities can be ordered through real-valued valuation functions like
price or size. As the real numbers constitute an ordered field, the analysis of such data benefits from the existence of the operators  $+, -, \cdot, /, 0, 1 $ together with total\footnote{Total refers to the property that for any two elements $a,b$ either $a\leq b$ or $b\leq a$ is true. A total order relation is often also called \emph{linear order}. } comparability $\leq$. Moreover, this combination  allows for measures of tendency (such as mean, variance, and skewness) and transformations. If more than one real-valued dimension is present, this yields to a real vector space $\mathbb{R}^n$, which results in additional descriptive measures and metric properties, such as volumes, angles, correlation, covariance. This is the standard setting for the majority of data analysis and machine learning models, and algorithms (\eg density-based clustering, logistic regression, SVMs, to name just a few).

\randbemerkung{Adequacy of numerical measurements}
However, organizing hierarchical relationships by means of numerical values is not always adequate, as this kind of organization presupposes two important conditions:
\begin{enumerate}
	\item %
	      every pair of entities has to be comparable, and
	\item %
	      the magnitudes of the differences between numerical values are meaningful and thus comparable themselves.
\end{enumerate}
In many situations, however, this is not the case: ($i$) does not hold, \eg in concept hierarchies (`mankind' is neither a subconcept nor a superconcept of `ocean') nor in organizations (a member of parliament is neither above nor below a Secretary of State), and also teaching curricula are far from being linearly ordered; ($ii$) does not hold, \eg in school grades (In the German system is the difference between `1 -- very good' and `2 -- good' equal to the difference between `4 -- sufficient' and `5 -- insufficient/fail'?) nor in organizations (In the European Commission, is an advisor closer to a deputy director general than a head of group to a director?).

\randbemerkung{Levels of measurements}%
\label{page-stevens}%
To address such variations of data types, S.\,S. Stevens has distinguished in~\cite{stevens1946theory} four \emph{levels of measurement: nominal, ordinal, interval,} and \emph{ratio}. For data on the \emph{ratio level} (\eg height), all above-mentioned operations are allowed (division, for instance, provides ratios). Data on the \emph{interval level} (\eg temperature measured in Celsius or Fahrenheit) do not have a meaningful zero as point of reference and thus do not allow for ratios, while the comparison of differences is still meaningful. Data on the \emph{ordinal level} (\eg the parent relation) can be compared hierarchically only, and on \emph{nominal level} (\eg eye color) only up to equality. Over time, this classification was discussed, refined and expanded, but never discarded~\cite{mosteller1977data,chrisman98}.

The mathematical field of Order and Lattice Theory \cite{harzheim2005ordered,caspard2012finite,schroder2016ordered,birkhoff1948lattice,Gratzer,davey2002introduction} is a canonical theory for modeling ordinal data. Its development started --- independently of the above-mentioned quest for a suitable theory for measurements --- in the second half of the 19th century as algebraic logic with the aim of George Boole \cite{boole1853investigation}, Charles S. Peirce \cite{peirce35collected} and Ernst Schröder \cite{schroeder90algebra} to formalize a calculus for concept hierarchies as cornerstone for a mathematical logic.
In symbolic knowledge representation, ordered sets are extensively used for this purpose. For instance, in RDFS, the set of all \texttt{rdfs:Class}es with the \texttt{rdfs:subClassOf} relation is an ordered set, and in the field of conceptual graphs, one frequently assumes that the type hierarchy is even a lattice \cite{sowa84conceptual}.

\section{Hierarchies in Data Science}
\label{sec-role-of-hierarchies}

Data Science is the scientific field of analyzing data and extracting knowledge from data. This understanding of data science as an interdisciplinary field grew over the past decades \cite{tukey1962future,piatetskyshapiro1991knowledge,escoufier1995science,davenport2007competing,dumontier2017science} as a confluence of methods from statistics, data mining/knowledge discovery, machine learning, data management and big data. Here, we briefly discuss the relationship between some of these areas to ordinal data.

In \fett{statistics}, several correlation measures exist for rankings (\eg Kendall's $\tau$ \cite{kendall1948correlation}, Spearman's $\rho$ \cite{spearman1904proof}, Goodman and Kruskal's $\gamma$ \cite{goodman79})
and tests for the significance of correlation (\eg Mann–Whitney U test and Wilcoxon signed-rank test  \cite{10.2307/3001968}). However, these measures and tests are designed only for \emph{linear orders}  (\ie multi-inheritance and pairs of incomparable elements are not considered).
In \fett{data mining and machine learning}, most methods are designed for numerical (and some for categorical) data. When ordered sets occur, then mostly as part of the model, \eg in decision trees \cite{breiman1984classification}, as topology of neural networks \cite{kleene1956representation}, as inclusion hierarchy for frequent item sets \cite{10.1145/170036.170072} (cf. next subsection), or as structure for coarser and finer set systems in hierarchical agglomerative clustering \cite{10.2307/2282967}. Often data points are organized in (linear) rankings, as a result from some numerical score, \eg the predicted accuracy in supervised learning, the cosine similarity to a given query in information retrieval, or the result of a learning-to-rank approach. Here again, these methods produce linear orders. A setting where the resulting order is non-linear is \fett{ontology learning} (see below).

Ordinal Data Science will profit from and contribute to the field of \fett{machine-learned ranking}~\cite{INR-016,DBLP:conf/icml/GouvertOF20}, particularly \fett{preference learning}~\cite{hullermeier2008label}, which is a subfield of AI concerned with label ranking and instance ranking. Other areas in AI do also turn to ordinal data, e.g.,  ``Deep Ordinal Reinforcement Learning''~\cite{ZapJF19} attempts to adapt reinforcement learning to ordinal data.
Machine learning methods for linear ordered sets are already somewhat further developed, e.g., treatment of linear ordered classes in supervised learning~\cite{cardoso2005modelling}. The same is true for the important machine learning method \emph{metric embedding}, which was recently adapted to linear ordered data~\cite{shi2016metric} and ordinal constraints inferred from metrics~\cite{pmlr-v35-kleindessner14,pmlr-v139-suzuki21a}.
An even more extensive study on the use of linear ordered data~\cite{keller2020ordinal} concluded that even in the presence of metric features the study of purely ordinal features might be fruitful. Early methods of how this might be done are already being explored in a rather practical way~\cite{ZhangC20a}. Finally, Ordinal Data Science is strongly related to the discipline \emph{statistical relational learning}, a research area that itself has a lot in common with many other AI areas, such as reasoning and knowledge representation. Foremost their advances in treating graph data~\cite{rossi2012transforming} is of high relevance to Ordinal Data Science and vice versa.

\section{Order-Theoretical Foundations}
\label{sec-foundations}
\subsection{Ordered Sets}
\label{sec:ordered-sets}
\randbemerkung{Poset}%
The key concept for studying ordinal data is an ordered set. An \fett{ordered set} $(P,\leq)$ is a directed graph (\ie it consists of a set $P$ and a binary relation $\leq$ on $P$) such that $\leq$ is reflexive ($p\leq p$), transitive ($p\leq q,\; q\leq r \implies p\leq r$), and antisymmetric ($p\leq q,\; q\leq p \implies p=q $) \cite{birkhoff1948lattice}.
This is equivalent to a transitive, directed acyclic graph with self-loops. Ordered sets could thus be seen as special cases of directed graphs or networks, and indeed some results of graph theory and network analysis are used; but order theory extends this with its own, specific flavor: For instance, the acyclicity of an ordered set simplifies its decomposition (\eg, by filters and ideals), results in more specific structural measures such as height, width and dimension, and allows for more readable visualisations. 

A typical ordered set is the set of all human beings together with the \texttt{is\_ancester\_of} relation. Another example are the dependencies of tasks in production planning. Note that, in general, we do not require comparability of all elements. The latter would mean that the order is \emph{total} (also called \fett{linear}), \ie that, for all $p,q\in P$, one of $p<q$, $p=q$ or $p>q$ holds. To emphasize that comparability is not required, ordered sets are also called \emph{partially ordered sets} (or \emph{posets} for short) in the literature, especially in computer science.

While one cannot exploit numerical operations for describing and analyzing ordinal data, they come with other constructions, such as \fett{order filters, order intervals, Pareto optima} etc.

For finite ordered sets --- as studied in data science --- we can reduce $\leq$ without loss of information to the \fett{neighboring relation} $\prec$ by removing all pairs that can be deduced by reflexivity and transitivity. This is the relation predominantly used for visualization. Note that in many cases, the order is not \fett{linear} (\ie there may be incomparable elements and multi-inheritance).

\label{what-are-lattices}
A particularly useful kind of order relations are \emph{lattices}, as they provide more structure and operations: In a \fett{lattice}, any two elements $p$ and $q$ always have a \emph{unique least common upper bound} $p\vee q$ and a \emph{unique greatest common lower bound} $p\wedge q$. These two operations are often called \emph{join} and \emph{meet} respectively. One could say that a lattice is a hierarchy in which one can navigate up and down with these two operations. A \emph{complete lattice} is an ordered set where every subset of $P$ (including the empty set and infinite subsets) has meet and join. This particular property is only relevant if $P$ is not finite, since each finite lattice is complete.
Although the datasets considered in data science are always finite, a deeper understanding of the infinite case, and hence of complete lattices, can provide new insights.

A complexity measure for an ordered set or lattice $(P,\leq)$ is its \fett{order dimension}, i.e.,  the number $n$ of linear orders $\leq_1,\dots, \leq_n$ on $P$ with $\bigcap_{i=1}^n\leq_i\,\,\,=\,\,\,\leq$. In other terms, it is the smallest $n\in\mathbb{N}$ such that the ordered set is embeddable in the Cartesian product of $n$ linear orders \cite{trotter2001combinatorics}. In a previous work~\cite{attributeselectionwithcontranominals} we explored how to reduce the complexity of a dataset by identifying parts that contribute to it having a high order dimension.
However, computing (i.e., deciding) the order dimension is \NPcomplete{} if the dimension is higher than two~\cite{Yannakakis}. Even approximations are proven to be \NPcomplete{}~\cite{HEGDE2007435}.

\subsection{Ordinal Structures}
\randbemerkung{Ordinal structure}%
As data may be ordered by different criteria, we define, more generally, an \fett{ordinal structure} $\mathbf{P}=\langle P, (\leq_i)_{i\in I}\rangle$ as a set $P$ of objects that is equipped with quasi-orders $\leq_i$~\cite{StrahringerWille92}.\footnote{The entities may -- and in most cases will -- have additional attributes on other levels of measurement. The development of hybrid analysis methods is of particular interest here.} \emph{Quasi-orders} are transitive, reflexive relations. Not insisting on anti-symmetry for $\leq_i$ means that there may exist objects that are indistinguishable in this relation (\eg there may be different objects with identical prize).

\begin{figure}
  \includegraphics[width=\textwidth]{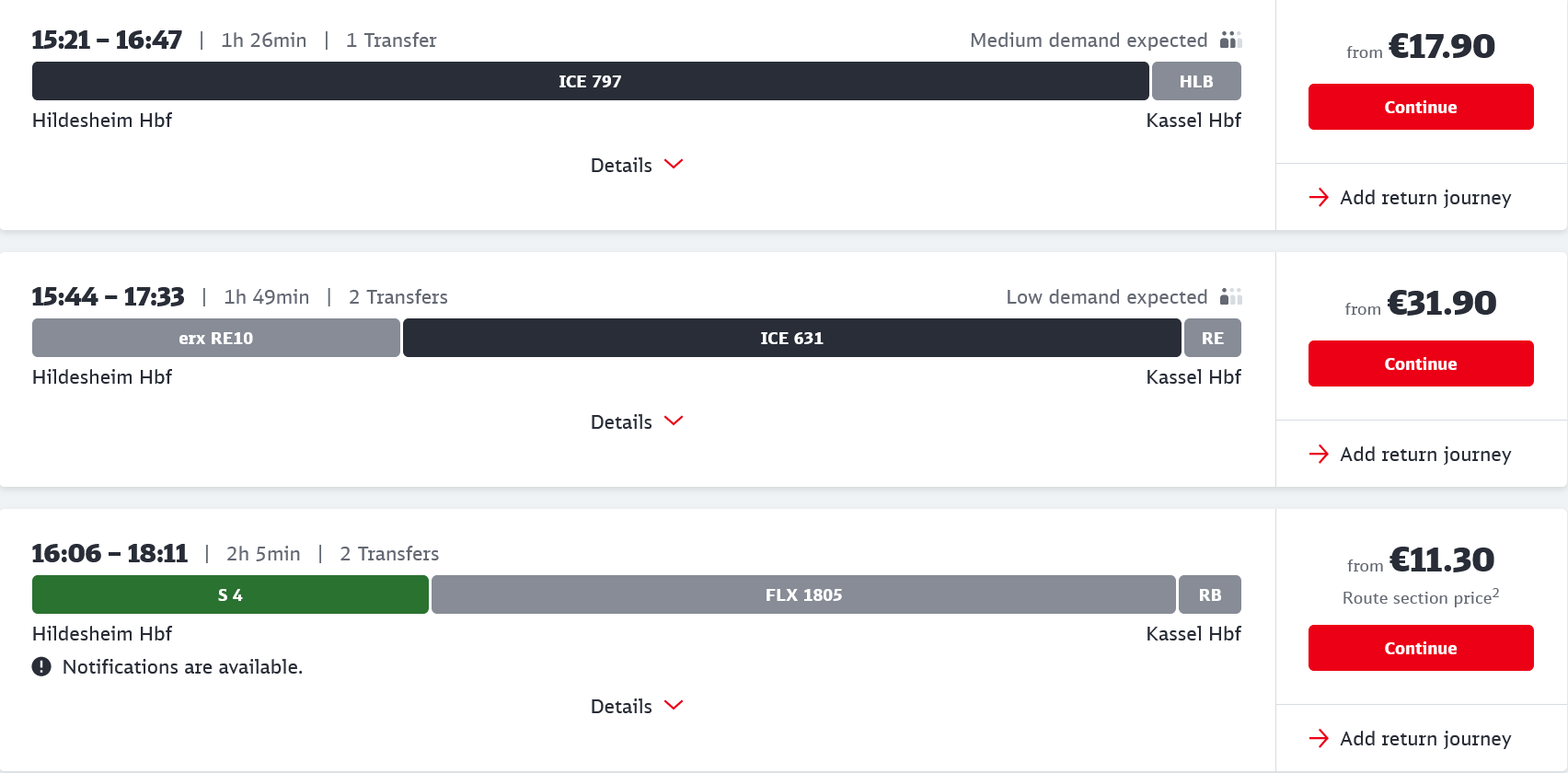}
  \caption{Pareto optima of train connections between the cities Hildesheim and Kassel as inferred from the German railway corporation Deutsche Bahn via \url{www.bahn.de}.}
  \label{fig-bahn}
\end{figure}

\begin{example}[Ordinal Structure]
	Train connections between two cities are an example for an ordinal structure. They can be ordered by different criteria, \eg by departure time (closer to time given in the query is better), travel duration (shorter is better), number of transfers (fewer is better), expected demand (lower is better), and price (lower is better). 	Figure~\ref{fig-bahn} shows the \emph{pareto optima} for connections between Hildesheim and Kassel for a requested departure at 3pm, \ie those connections for which it is impossible to improve a criterion without becoming worse in any of the other criteria.
\hfill\qed
       
\end{example}

In our vision, we extend the notion of ordinal scale to allow for the analysis of all kinds of ordinal  data. With terms like `\emph{order}', `\emph{ordinal}', we always refer to arbitrary orders --- be they linear or not.\footnote{ When addressing linear orders, we will state this explicitly. For sake of linguistic simplicity, we will also summaries all levels of measurement above ordinal --- essentially interval and ratio --- as `\fett{numerical}'.} We thus go beyond  Stevens' original understanding of \fett{`ordinal scale'} which only referred to linear orders. Even though this seems to be only a minor modification at first glance, it has far-reaching consequences for the question of what  `measurement' actually is, as well as for the complexity of computational problems.

Ordered sets in general do not provide much structure. However, every ordered set is embeddable into a complete lattice, its \fett{Dedekind-MacNeille completion}~\cite{zbMATH03027979} (which generalizes Dedekind's  construction for embedding $\mathbb{Q}$ in $\mathbb{R}$). This allows for tapping into the rich algebraic theory of lattices and of universal algebra~\cite{birkhoff1946universal,gratzer1968universal,burris1981course}. The class of lattices forms a so-called \fett{variety}, which implies that every homomorphic image, every factor lattice and every subset of a lattice that is closed under join and meet, as well as all direct products and direct sub-products of lattices are lattices again. Hence, these operators can be exploited for composing and decomposing lattices --- and thus eventually of the ordered sets that generated them. In the worst case, the size of the Dedekind-MacNeille completion is exponential in the size of the ordered set. However, this growth is expected to be tame for real-world data, due to low amounts of multi-inheritance in it. Ganter and Kuznetsov provided an algorithm for its computation~\cite{10.1007/BFb0054922} in O$(c\cdot w \cdot n^2)$ with $c$ being the size of the completion, $w$ the width of the ordered set and $n$ its size. The new nodes bear (yet unexplored) potential for identifying substructures in ordered sets.

\subsection{Formal Concept Analysis and knowledge spaces}
\randbemerkung{Formal Concept Analysis}
\label{page-fca}
A direct application of lattice theory to data analysis has been established as \fett{Formal Concept Analysis} (FCA) by R. Wille~\cite{wille82restructuring}. It follows the spirit of Boole, Peirce and Schröder by computing concept hierarchies from datasets. In its most basic form, FCA derives a \fett{concept lattice} from a so-called \fett{formal context}, a dataset $(G,M,I)$ consisting of a set of objects $G$, a set of binary attributes $M$, and $I\subseteq G\times M$ indicating which objects have which attributes. \fett{Concepts} are all maximal pairs $(A,B)$ with $A\subseteq G$ and $B\subseteq M$ s.\,t.\ all objects in the \fett{extent} $A$ share all attributes in the \fett{intent} $B$, and vice versa. The set of formal concepts is denoted by $\BV(G,M,I)$ and the relation ${\leq}\subseteq\BV(G,M,I)\times \BV(G,M,I)$ with $(A,B)\leq(C,D)\longeq A\subseteq C$ constitutes an order relation on $\BV(G,M,I)$. Thereby $(A,B)$ is designated as a subconcept of $(C,D)$. More importantly, $\leq$ is a lattice order. This fact follows from the \emph{basic theorem on Concept Lattices}~\cite[Theorem 3]{ganter96formale}. This theorem states that for any subsset $\{(A_{i},B_{i})\}_{i\in I}$ of $\BV(G,M,I)$ there exists an infimum (meet) $\bigwedge_{i\in I} (A_{i},B_{i})$ and a supremum (join) $\bigvee_{i\in I}(A_{i},B_{i})$ in $\leq$. The set of all concept extents forms a closure system, \ie it is a set system which is closed under intersection, and the same holds for the set of all concept intents. 

\begin{figure}[t]
  \centering
\includegraphics[width=8cm]{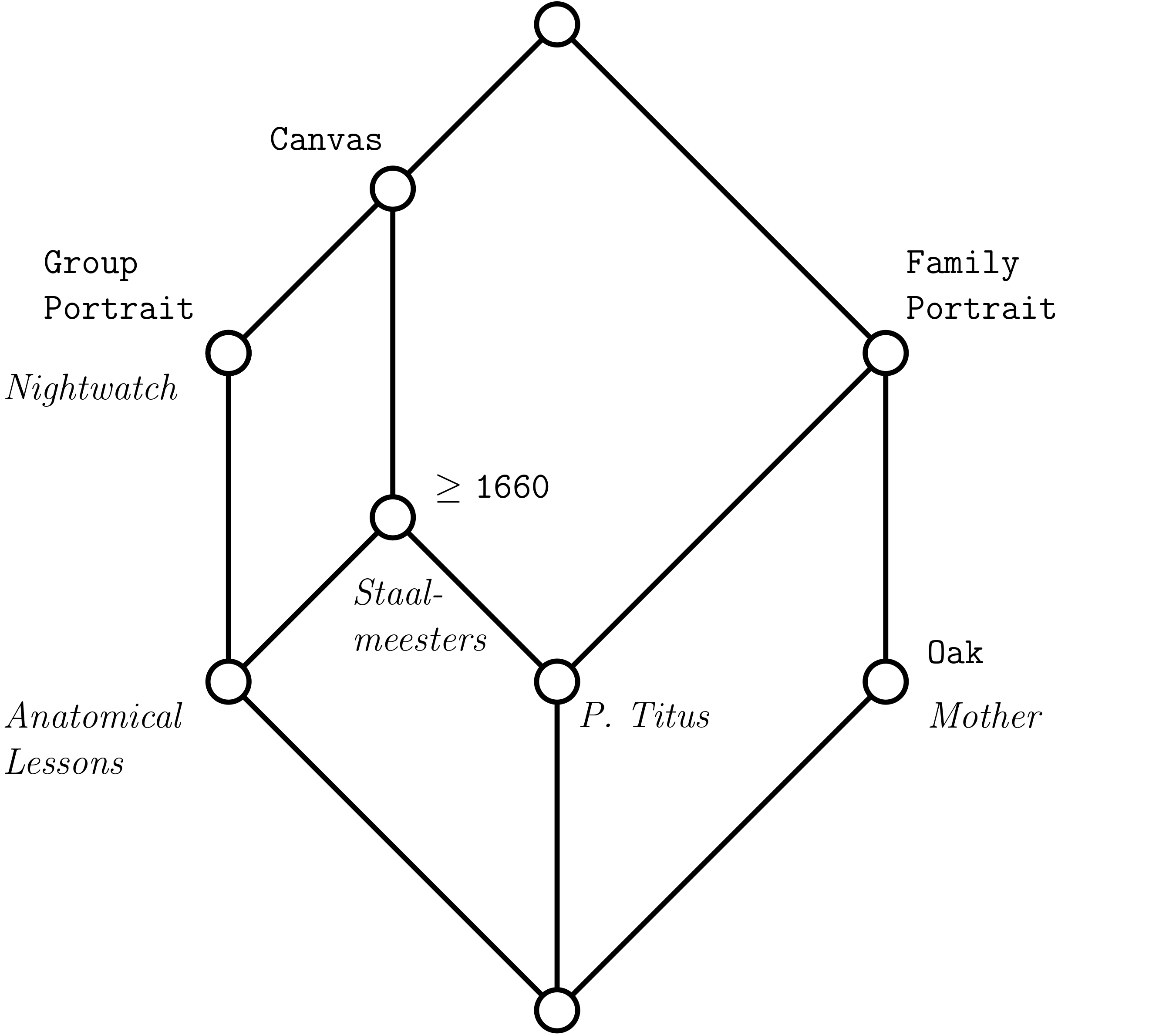}
  \caption{Line diagram of the concept lattice from  Example~\ref{ex:fca}. The drawing uses a short-hand notation for the labels. The nodes represent the set of formal concepts.  Attributes are drawn on top of nodes and objects below. Only the first occurrence of an attribute and the last occurrence of an object is annotated (in the reading direction from top to bottom).}
  \label{fig:remlat}
\end{figure}

\begin{example}
  \label{ex:fca}
  The following table depicts a real world data set about paintings by
  Rembrandt (the object set $G$) and their properties ($M$), taken
  from ``Concept lattices and conceptual knowledge
  systems''~\cite{Wille92}. The formal context $\context=(G,M,I)$ and
  its incidence $I$ is represented by means of a \emph{cross table}:
  
  \begin{center}
      {\small
        \begin{cxt}
          \cxtName{$\context$}
          \att{Family Portrait}
          \att{Group Portrait}
          \att{Oak}
          \att{Canvas}
          \att{$\geq$ 1660}
          \obj{.x.x.}{\textit{Nightwatch}}
          \obj{.x.xx}{\textit{Anatomical lessons}}
          \obj{x..xx}{\textit{Portrait Titus}}
          \obj{...xx}{\textit{Staalmeesters}}
          \obj{x.x..}{\,\textit{Mother}}
        \end{cxt}}
  \end{center}

  The line diagram of the concept lattice corresponding to $\mathbb{K}$ is shown in Figure~\ref{fig:remlat}. Such a line diagram displays the covering relation, i.e., the subset of $\leq$ which contains only the comparable elements that are immediate neighbors. From the lattice we can infer various types of information. For example, we find that all paintings that were painted in 1660 or later were painted on canvas. More intriguingly, family portraits that were painted on canvas occurred in 1660 or later. For a comprehensive list of how-to infer information from lattices, and in particular lattice diagrams, we refer the reader to the literature~\cite{ganter96formale}. For a more in-depth discussion of questions concerning the diagrammatic representation of order structures in the specific case of the present work, please refer to~Section~\ref{sec-exploration}.\hfill\qed
\end{example}

Via \fett{conceptual scaling} \cite{GanterWille89}, FCA can be applied to any kind of data. 
Since its foundation in 1982, FCA has grown to an extensive theory \cite{ganter99formal,carpineto04concept,books/sp/GanterO16,Ferre2020} and has seen several extensions, for instance, to fuzzy concepts \cite{belohlavek2012fuzzy}. It has been applied in a large variety of domains (see, e.g., \selfcite{stumme00begriffliche,ganter05formal} for surveys). FCA has been connected to other disciplines, such as Software Engineering \cite{snelting00understanding}, Psychology \cite{spangenberg1988conceptual}, Scientometrics \selfcite{doerfel2012publication},
and Social Choice Theory \cite{10.1007/978-3-540-39644-4_17}.

Similar to FCA is the theory of \emph{knowledge spaces} by J.--P. Doignon and J.--C. Falmagne \cite{Doignon_Falmagne_1999}. They define a \emph{knowledge structure} as a pair $(Q,\mathcal{K})$ where $Q$ is a non-empty set and $\mathcal{K}$ is a family of subsets of $Q$. The set $Q$ is called \emph{domain} and its elements are called \emph{questions} or \emph{items}. The subsets in $\mathcal{K}$ are called \emph{(knowledge) states}. The intuition for this definition is that $Q$ is a collection of questions of some school subject, and that the sets in $\mathcal{K}$ describe which combinations of questions could be solved by individual students. A \emph{knowledge space} is then a knowledge structure that is closed under set union. This is inspired by the assumption that if two students have knowledge states $K_1$ and $K_2$, resp., then it should be possible for them to learn what the other student knows, resulting in $K_1\cup K_2$. 

A knowledge space is thus a kernel system, \ie a set system closed under union. If $\mathcal{K}$ is a kernel system, then $\{Q\setminus K \mid K\in\mathcal{K}\}$ is a closure system, and vice versa. This means in particular that all theoretical and algorithmic results of FCA can directly be transferred to knowledge spaces, and vice versa (see in particular \cite{DBLP:conf/icfca/GanterG14}). From a mathematical viewpoint, the structures of closure systems, kernel systems, complete lattices and concept lattices are equivalent~\cite{wille82restructuring}, as there exist natural 1-to-1 correspondences. These structures are also in a natural 1-to-1 correspondence to sets of implications (of propositional Horn logic)~\cite{wild1994theory} and functional dependencies (of database theory). 
These natural correspondences mean that for the computation, analysis and (de)compositions of such structures one can switch to the representation that is most suitable for the task at hand. Formal contexts play an important role when it comes to computations, as their sizes are logarithmic in comparison to the sizes of the  corresponding (concept) lattices.

\subsection{Ordered Sets and Lattices in Computer Science}
\randbemerkung{Posets in computer science}%
Specific ordered sets and lattices are frequently found in computer science: For instance, queues and lists are linear orders, (B-)trees are used for indexing data, tries for Information Retrieval, CPOs for modeling the semantics of recursion, and Boolean algebras for propositional logic implemented in logical gates. Inheritance (including multiple inheritance) in object-oriented programming induces a partial order on the set of classes. An introduction to the mathematical foundations of order theory relevant to \fett{Data Mining} is provided in~\cite{simovici2008mathematical}.
Ordered sets are also the key ingredient for modeling dependencies between tasks in scheduling problems.

FCA is used extensively in the field of \fett{Ontology Engineering\,/\,Semantic Web}, in particular for conceptual information systems \selfcite{Stumme1999acquiring}, knowledge acquisition \selfcite{stumme1995knowledge,Stumme96a,Stumme97a}, for knowledge base completion in Description Logics \cite{baader95computing,baader07completing}, for computing probably approximately correct implication bases \groupcite{conf/icfca/BorchmannHO17}, for ontology management \selfcite{stumme01bottom,ganter03creation,cimiano2004concept}, \cite{roth2006latticebased,Bendaoud2008FormalCA}, and text corpus analysis \selfcite{hotho03explaining}.
An extensive survey on FCA based models and techniques in knowledge processing is given in \cite{DBLP:journals/eswa/PoelmansKID13}.

\subsection{Algorithms and Software for Ordered Sets and Lattices}
The number of algorithms and software tools in the field of order theory and (concept) lattices is extensive. Listing all relevant works would go far beyond the scope of this article; we only mention the most important and recent tools. Ordinal Data Science may benefit from algorithms for the uniform sampling of linear extensions~\cite{bubley1999faster} and the uniform sampling of ordered sets~\cite{KOZIEL2020294}, and the even more complicated problem for sampling binary relations in general, which was tackled in \groupcite{conf/iccs/FeldeH19}.
The research field of Formal Concept Analysis can contribute algorithms for the fast generation of concept lattices, e.g.~In-Close4~\cite{andrews2017making} or Parallel Close-by-One~\cite{krajca2010parallel}. Also, the computation of (inter-feature-) dependencies will be an essential asset. The computation of such dependencies can be done with proven parallel canonical base algorithms~\cite{kriegel2017nextclosures}. One may also resort to approximate methods, since the related computational problems become computationally intractable with increasing input size. For example, the dependencies can be computed with \emph{probably approximately correct learning} methods~\groupcite{BORCHMANN202030,conf/icfca/BorchmannHO17}, a combination of order theory and classical machine learning model.
Many of the mentioned algorithms are already available, or are currently being implemented, in the analysis software \texttt{conexp-clj}~\groupcite{conf/icfca/HanikaH19}, which is a reimplementation of \texttt{conexp} by means of the functional programming language \emph{Clojure}. It is currently maintained and further developed in the group of the authors. There are many further FCA software tools under active developed (and even more historic ones), most with a particular specialization. Just to name a few: \emph{Graph-FCA}\footnote{\url{https://bitbucket.org/sebferre/graph-fca/src/master/}}, an extension of FCA to knowledge graphs; \emph{lattice miner}~\cite{missaoui2017lattice}; \emph{conexp-ng}\footnote{\url{https://github.com/fcatools/conexp-ng}}, an educational variant of \emph{conexp}; \emph{fcaR}\footnote{\url{https://cran.r-project.org/web/packages/fcaR/fcaR.pdf}}, a package for the statistical software suite R.

\bigskip
\noindent
In the next five sections, we will now discuss exemplarily five out of many possible lines of research in the new field of Ordinal Data Science in more detail.

\section{Ordinal Measurement Theory}
\label{sec-ordinal-measurement-theory}

While measurements in physics were accepted without objections, other disciplines, in particular psychology, asked in the 20th century --- and still do so today --- if and under which conditions non-physical sensations such as loudness, fear, or intelligence can be measured.
The most widely acknowledged answer to this is the \fett{Representational Theory of Measurement (RTM)} \cite{krantz1971foundations,suppes1989foundations,luce1990foundations}, which aims at identifying criteria that justify a measurement on a given level. These criteria can (at least in principle) be verified empirically without actually performing the measurement. RTM defines \fett{measurement} as a homomorphism from an empirical relational structure $\mathbf{A}=\langle A,(R_i)_{i\in I}\rangle$ --- \ie a set $A$ of empirical objects (such as a collection of rods) equipped with qualitative relations $R_i$ (such as `rod $a$ is shorter than rod $b$' and `when laid end to end, rods $a$ and $b$ together have the same extension as rod $c$') --- to a \emph{numerical relational structure} $\mathbf{B}=\langle B,(S_i)_{i\in I}\rangle$ \cite{pfanzagl1971theory,krantz1971foundations}. In the latter, $B$ is usually the set $\mathbb{R}$ of real numbers (or $\mathbb{R}^n$ and other variations) together with relations such as $x<y$ and $x+y=z$, even though several researchers \cite[p. 253]{roberts1984measurement} observe that in some cases it might be beneficial to allow for other algebraic --- `non-numerical' --- structures.

\randbemerkung{Conceptual measurement}%
RTM puts a strong emphasis on measurements into the real numbers, and less emphasis on ordinal phenomena. As it allows for linear ordinal scales only, RTM does not allow for studying phenomena involving incomparability or multi-inheritance.
Therefore, inspired by RTM, R. Wille et al.~\cite{ganter86conceptual} developed a theory of \emph{conceptual measurement} where $\mathbf{B}$ takes the form of specific lattices, called \fett{conceptual scales} --- a first step towards an ordinal version of RTM. In this context several studies followed \cite{GanterWille89,RUWille96,journals/dam/PollandtW05,stumme96local,stumme99hierarchies}. Conceptual measurement provides structural statements about homomorphisms to standard scales,
but it does not yet provide a full theory about the conditions under which ordinal phenomena can be measured with the different types of ordinal scales.
In order to provide a \fett{theoretical foundation to Ordinal Data Science}, we propose to extend RTM by non-numerical scales. This will allow for empirically testing on a given dataset which scales are appropriate for its analysis.

To this end, in this section, we suggest to analyze under which conditions measurements into (linear and in particular in non-linear) ordinal scales exist, and how unique they are (\ie under which permissible transformations they remain invariant). We further suggest investigating how the --- potentially high --- order dimension of the original data relates to the order dimensions of a set of scales which together fully measure the dataset. At last, we consider non-linear versions of Guttman scales~\cite{guttman1944basis}, to allow for measuring simultaneously objects and their features.

In the long run, this endeavor may open the avenue to a
\fett{`Grand Unified Theory of Measurement'} which will allow, for instance, in (the philosophy of) physics to discuss the analysis of symmetries of elementary particles  (by letting $\mathbf{B}$ be a (non-ordered) symmetry group) in the same terms as the numerical measurement of their mass and charge. %

\subsection{State of the Art and Open Questions}

Based on the successful role of measurement in physics and the demand of other disciplines --- in particular psychology ---, in the late 19th and early 20th century a universal definition of measurement was sought \cite{ferguson1940quantitative}.
\randbemerkung{Scale types}%
S.\,S. Stevens \cite{stevens1946theory} provided an operational approach by distinguishing four \fett{levels of measurement}: nominal, ordinal, interval, and ratio; and characterized them by families of automorphisms of the set of real numbers that respect the scale type (`admissible functions').
Stevens' levels of measurement have been (and still are \cite{scholten2009reanalysis,azu_jmmss23785}) under dispute. A particular controversial question among practitioners that is discussed since Stevens' paper for over 70 years is whether computing the mean of ordinal data is an allowed operation or not (see, \eg the football number dispute \cite{lord1953statistical,scholten2009reanalysis} as illustrative example). In practice, this is frequently done (\eg  aggregating the jury votes in gymnastics or figure skating), while in other cases this is considered bad practice (\eg aggregating reviewer judgments in scientific peer review, where often the average is considered as bearing little meaning, and therefore consensus is sought). Several extensions were proposed for Stevens' levels  (\eg \cite{mosteller1977data,chrisman98}), see \cite{sep-measurement-science} for a systematic survey. All those extensions, however, consider ordinal data as a separate category; so our endeavor is widely agnostic to this discussion.

\randbemerkung{Conjoint Measurement}%
Stevens did not answer the question of how to decide, in a given setting, which level is appropriate. \fett{Representational Theory of Measurement (RTM)} \cite{krantz1971foundations,suppes1989foundations,luce1990foundations}, \cite{pfanzagl1971theory,roberts1984measurement} provides an answer with the notion of conjoint measurement.  RTM  defines a \emph{measurement} as a mapping of an empirical relational structure (ERS), \ie a set of objects equipped with qualitative relations such as `is larger than', to a numerical relational structure (NRS). Luce and Tukey \cite{Luce64simultaneousconjoint} proved that whether some empirical sensation is a continuous variable depends on characteristics of the ERS and is thus in principle empirically testable. How feasible this testing is in practice is one of the questions of the still ongoing discussion on RTM, in particular in psychology \cite{trendler09measurement,doi:10.1177/0959354315617253}. The envisaged substantially new methods in Ordinal Data Science for measuring ordinal data may turn RTM more acceptable to practitioners, as it is easier to show that ordinal methods are legible in a given situation than it is for methods requiring interval or even ratio level.

Several researchers working on RTM (\eg \cite[p. 253]{roberts1984measurement}) stated that it might be beneficial not to focus solely on the ordered field of real numbers as NRS, but to allow instead for other mathematical structures. However, this approach has been followed rarely. For instance, R. Wille and his group took up this line of thought, as described above:
\randbemerkung{Conceptual measurement}%
On one hand the group studied the representation of ordinal structures in the real numbers \cite{RUWille95,RUWille96,wille2000linear} and weaker algebras such as ordered quasi-groups \cite{DBLP:journals/ejc/Wille96}. On the other hand, work on a theory of \emph{conceptual measurement} \cite{ganter86conceptual,GanterWille89,stumme96local,stumme99hierarchies,journals/dam/PollandtW05}  was started,
\randbemerkung{Partial ontologies}%
where $\mathbb{R}$ is replaced by complete lattices which can be understood as \emph{partial ontologies}.

As Dedekind-MacNeille completions preserve important properties of ordered sets (\eg order dimension) and are isomorphic to a concept lattice, conceptual measurement is just a special --- but conceptually and computationally convenient --- way of formalizing a general theory of ordinal measurement.  Conceptual measurement has frequently been used in applications. However, a theoretical investigation in terms of RTM (\eg what are the equivalences of cancellation and solvability when $\mathbb{R}$ is replaced by a lattice?) is still lacking.

\subsection{Promising Research Questions}

The main objective in this field of research will be to \fett{extend Relational Measurement Theory to non-numerical scales}, as it would allow to empirically test on given data which of these scales are appropriate. We will address three key aspects: The first is to establish \fett{theorems about the existence and uniqueness of measurements into non-linear ordinal scales} and to explore how well the \fett{theory can be extended to fundamentally different scale types} (\eg (non-ordered) symmetry groups), as a first step towards a \fett{`Grand Unified Theory of Measurement'}.
As the order dimension is a strong indicator for the complexity of an ordered set, we will then discuss the \fett{relationship between the order dimensions} of an ordinal structure and its scales, in order to provide guidance for suitable decompositions and factorizations in Sections \ref{sec-algebraic-constructions} and \ref{sec-ordinal-factor-analysis}; %
and extend the discussion to scales that measure simultaneously the objects and their attributes.

\subsubsection{Measuring into ordinal scales}
Research in Representational Measurement Theory is centered around the real numbers as `numerical relational structure' (including derived structures, such as semiorders and interval orders for modeling error and variation) and real vector spaces.
\randbemerkung{Measuring into arbitrary ordinal scales}%
Measurements with other types of ordered sets have not been studied systematically, except for the theory of conceptual measurement as described above. So a general research question will be: how does measurement theory look like if we replace $(\mathbb{R},\leq,+,-,\cdot, 0,1)$ by a specific ordered set $(P,\leq)$ or (complete) lattice $(L,\leq,\wedge,\vee)$? Are there equivalences to the notions of cancellation, solvability and conjointness, and how are they constituted?

\randbemerkung{Representation problem}%
Following the line of traditional RTM, we suggest to study in particular the representation problem and the uniqueness problem for specific scale types. The \emph{representation problem} deals with the question of which conditions a dataset has to fulfil in order to permit a measurement in a given scale. For binary attributes on the empirical side and many standard ordinal scales (nominal, linear, interval, multi-ordinal, contra-nominal, contra-ordinal, and convex-ordinal scales) on the `numerical' side, this problem was addressed in \cite[Chapt.~7]{ganter99formal}. In this case, the task is reduced to translating the results to the terminology of RTM. For other types of empirical data, and for other types of scales (\eg trees) and ordinal scales in general\footnote{Reminder: In this project, ordinal scales may be any type of ordered set, either linear or non-linear. What is usually called ordinal scale (such as Likert scale or Beaufort) is a special, linear case thereof.} on the `numerical' side, the necessary theorems still have to be established.
\randbemerkung{Uniqueness problem}%
Once a measure has been identified, one should also consider the \emph{uniqueness problem} by analyzing invariance under permissible transformations: How unique is the resulting measure or scale?

Our results constitute the first steps towards a \fett{`Grand Unified Theory of Measurement'}:
Once a theory for ordinal measurement as been established, one may start to explore --- as a side-trip and preparation of more extensive research --- how well this theory can be extended to fundamentally different scales, as for instance to non-ordered groups (such as symmetry groups).

\subsubsection{Dimensionality}
The complexity of an ordered set is not only determined by its cardinality, but also to a large extent by its order dimension. In contrast to numerical data, where most frequently each feature spans one dimension, the identification of the different dimensions of an ordered set is not straightforward --- even determining their number is \NPcomplete~\cite{Yannakakis}. The question of dimensionality thus takes a completely different stance as in standard RTM (where dimensional analysis means that physical dimensions such as length, mass, time duration, speed, and their relationships are studied \cite[Chapt. 10]{krantz1971foundations}).
\randbemerkung{Dimensional analysis of collections of scales}%
Breaking down a complex ordered set into smaller, less complex parts (\ie parts with lower dimensionality) is an important --- but challenging --- task for data science, which we will discuss further in Sections \ref{sec-algebraic-constructions} and \ref{sec-ordinal-factor-analysis}. The following example shows that
the development of a theory about the dependencies between the ordinal dimension of an ordered set and the dimensions of the scales it can be measured to might be of high interest.

\begin{table}[t]
	\centering
	\caption{The final league table of the 2022–23 Bundesliga, Germany's premier soccer league. (Pos: Position, W: Won, D: Drawn, L: Lost, GF: Goals for, GA: Goals against, GD: Goal difference, Pts: Points)}
	\label{bundesliga}
	\begin{tabular}{llrrrrrrrr}
		\toprule
		\textbf{Pos} & \textbf{Team}            & \textbf{W} & \textbf{D} & \textbf{L} & \textbf{GF} & \textbf{GA} & \textbf{GD} & \textbf{Pts} \\
		\midrule
		1            & FC Bayern München        & 21         & 8          & 5          & 92          & 38          & +54         & \textbf{71}  \\
		2            & Borussia Dortmund        & 22         & 5          & 7          & 83          & 44          & +39         & \textbf{71}  \\
		3            & RB Leipzig               & 20         & 6          & 8          & 64          & 41          & +23         & \textbf{66}  \\
		4            & 1.\ FC Union Berlin      & 18         & 8          & 8          & 51          & 38          & +13         & \textbf{62}  \\
		5            & SC Freiburg              & 17         & 8          & 9          & 51          & 44          & +7          & \textbf{59}  \\
		6            & Bayer 04 Leverkusen      & 14         & 8          & 12         & 57          & 49          & +8          & \textbf{50}  \\
		7            & Eintracht Frankfurt      & 13         & 11         & 10         & 58          & 52          & +6          & \textbf{50}  \\
		8            & VfL Wolfsburg            & 13         & 10         & 11         & 57          & 48          & +9          & \textbf{49}  \\
		9            & 1.\ FSV Mainz 05         & 12         & 10         & 12         & 54          & 55          & $-$1        & \textbf{46}  \\
		10           & Borussia Mönchengladbach & 11         & 10         & 13         & 52          & 55          & $-$3        & \textbf{43}  \\
		11           & 1.\ FC Köln              & 10         & 12         & 12         & 49          & 54          & $-$5        & \textbf{42}  \\
		12           & TSG 1899 Hoffenheim      & 10         & 6          & 18         & 48          & 57          & $-$9        & \textbf{36}  \\
		13           & Werder Bremen            & 10         & 6          & 18         & 51          & 64          & $-$13       & \textbf{36}  \\
		14           & VfL Bochum               & 10         & 5          & 19         & 40          & 72          & $-$32       & \textbf{35}  \\
		15           & FC Augsburg              & 9          & 7          & 18         & 42          & 63          & $-$21       & \textbf{34}  \\
		16           & VfB Stuttgart            & 7          & 12         & 15         & 45          & 57          & $-$12       & \textbf{33}  \\
		17           & FC Schalke 04            & 7          & 10         & 17         & 35          & 71          & $-$36       & \textbf{31}  \\
		18           & Hertha BSC               & 7          & 8          & 19         & 42          & 69          & $-$27       & \textbf{29}  \\
		\bottomrule
	\end{tabular}
\end{table}

\begin{example}[Order Dimension]
	Table \ref{bundesliga} depicts the final standings of the German Soccer Bundesliga season in 2022/23.
	In this dataset, the strength of the soccer clubs is measured using four different scales.
	In the columns “Won” and “Goals for”, a high value implies a strong club, while the converse is true for “Lost” and “Goals against”.
	The direct product of these four linear orders describes an order relation in which two clubs are comparable, if one dominates the other in all four categories. Two clubs may be incomparable in this ordering, as for instance Borussia Dortmund has more wins than FC Bayern München, but also more losses.

	The order dimension of the domination order of the 2022/23 season is three. This means that only three linear orders are necessary to represent the combination of the four scales. There is hence some redundancy in the data --- but not enough to produce a total ranking. The latter would hold if the order dimension of the domination order would be one. It would then, for every pair of clubs, provide an answer to the natural question of which of the two is better than the other. As this is not the case, further rules are used to compute the final position of each club in the league. This results in a linear extension of this domination order. The choice of these rules captures the understanding which quality criteria are considered most important.

	The current rules of Bundesliga \cite{bundesliga} are one possible way to reach such a linear extension:
	First, a “Points” scale is derived, where each team is awarded three Points for every win and one point for every draw.
	This results in a linear extension of the direct product of the “Won” and “Lost” scales, which are considered to be the most important criteria.
	However, this may not yet result in a total ranking as multiple teams can have the same number of points (for instance Borussia Dortmund and Bayern München).
	In this case, the tie is broken using the “Goal difference”, which itself is a linear extension of the direct product of “Goals for” and “Goals against”.
	Thus, from the original four scales, the two linear extensions “Points” and “Goal difference” are computed as derived scales.
	Then, these two give the final ranking by taking once again a linear extension of the two derived scales.\footnote{\cite{bundesliga} provides further rules for cases where this still does not resolve all ties.}

	Note that the rules are arbitrarily chosen (\eg, one could assign only two points for a win, as is done in other sports (and was done earlier in Bundesliga), or consider the direct comparison of clubs in a tie as second criterion) and for every linear extension of the domination order there is a set of rules that gives rise to the ranking described by this extension. Thus, one could consider each linear extension a valid final ranking of the participating teams.\qed

\end{example}

A next research question could be a dimensional analysis of combinations of the standard ordinal scales --- for each of them studied separately, the answer is straightforward, but dimensional properties of full measures of an ordered set in a (heterogeneous) collection of scales have not yet been studied systematically.
As deciding the order dimension of an ordered set is \NPcomplete{}, one has to keep an eye on the computational tractability: Are exact algorithms fast enough for typically sized datasets? Alternatively one may resort to computing upper bounds, together with the \emph{realizers} (\ie the spanning linear orders) \cite{DBLP:journals/jal/YanezM99}. Also, the development of efficient algorithms for computing lower bounds (and eventually also the critical pairs) will be of interest.

\subsubsection{Joint scales on objects and attributes}
\emph{Guttman scales} have been introduced as joint scales for objects and attributes. Consider, for instance, a set $G$ of students, a set $M$ of exercises of (assumingly) increasing difficulty, and a relation $I$ stating which student solved which exercise. If the corresponding concept lattice is linearly ordered then it provides a linear ranking both for the difficulty of the exercises and for the ability of the students. This can be interpreted such that there is no exception to the rule: a student who is capable of solving a specific exercise can also solve all easier exercises. This is equivalent to the original definition  \cite{guttman1944basis} which requires the existence of maps $s\colon G\to\mathbb{R}$ and $e\colon M\to\mathbb{R}$ s.\,t. $(g,m)\in I \iff s(g)\leq e(m)$. In practice, Guttman scales can be observed frequently, but often only up to some `noise' or `impurity'. Standard approaches are then either to assume that the assumptions hold in principle and to statistically find best-fitting functions \cite{coombs1978coombs}, or to analyze whether the data result from more than one Guttman scale. We propose to follow two new order-theoretic approaches for analyzing these `impurities' instead.
\randbemerkung{Generalized Guttman scales}%
The first approach is to relax the condition of linear order and to explore if a more general scale --- \eg a tree --- may be suitable as a joint scale. One can then identify necessary and sufficient conditions for the existence of such generalized Guttman scales as well as criteria for their minimality (\eg the lowest cardinality, the lowest dimensionality, least multi-inheritance).
\randbemerkung{Scales with local impurity}%
The second approach is to analyze more closely the disturbance of the linear scale by identifying which parts of the scale increase its \emph{local dimension} \cite{journals/endm/TrotterW17}. Such a tool would allow analysts to investigate whether a linear feature is disturbed by `noisy data' or whether it really shows some unexpected non-linear characteristics.

\section{Ordered Metric Spaces}
\label{sec-metric-ordered-spaces}

Real-world data is feature-heterogeneous~\cite{cai2013multi}, i.e., it is (apart from others) composed of ordinal and numerical dimensions. The former can be viewed as a partially ordered set, and the latter often leads naturally to a metric. In principle, both can be analyzed separately, but this approach has two disadvantages. First, the results calculated in this way can be contradictory to each other. Second, the knowledge about the connection of the two feature dimensions is not exploited.

We therefore propose to develop theoretical and practical methods that honor the compatibility (or consistency) of order relations and  metrics on a set. To this end, we envision the unified representation of such data by means of  \fett{ordered metric spaces (om-spaces)} $(P,R,d)$, where $P$ denotes a set, $R$ is an order relation on $P$, and a $d$ a metric $d\colon P\times P\to \mathbb{R}^+$. This modeling is inspired by similar investigations in \emph{metric measure spaces}\footnote{In metric measure spaces, \emph{measures} are those of mathematical analysis, being defined over $\sigma$-algebras~\cite{halmos2013measure}. They model a different aspect of the  concept of measurement as the homomorphisms between relational structures in RTM do, and should not be confused with them.}~\cite{memoli2011metric,sturm2006geometry}.

The corresponding highly ambitious task, due to its novelty, is to develop a \fett{structure theory for om-spaces}. Particularly interesting in this context is the study of om-space  embeddings  into Euclidean space. Since the latter is a very commonly used space for machine learning, the question of how to measure \emph{ordinal distortions} is imperative. Vice versa, it is important to understand to which extent the extraction of linear orders from data via valuation functions is meaningful with respect to om-spaces. With a new class of distortion measures --- and computationally tractable approximation algorithms ---  which allow to \fett{quantify the compatibility of $R$ and $d$}, one might be able to answer a range of fundamental questions about the relationship between orders and metrics, including: 1) If a metric has been (machine) learned based on information about the order, how consistent is the metric with the order? This will provide a fundamental new quality criterion for order-related machine learning tasks. 2) How strong do order and metrics ``agree'' with their view on the dataset? An answer might be highly beneficial to an investigation whether order and metrics reflect the same hidden structure. 3) How justified is it to represent the order by means of a metric? If the answer is satisfactory for a given dataset, this would justify the use of the rich toolkit of numerical data analysis and machine learning methods for analyzing the ordinal structure. If not, this would indicate that alternative (ordinal) methods are mandatory in this particular case.

\subsection{State of the Art and Open Questions}
Order relations entail many geometric properties. Likewise, metric spaces, which play a dominant role in data science, have a rich geometric structure.
Examples for datasets exhibiting both a metric and an order relation are numerous. For example: Humans have (distances between their) birthplaces and an ancestry order relation; scientists have an ancestry relation (doctoral advisor) and different proximities (topical, geographic, social); mountains are ranked by height and prominence \selfcite{schmidt2018prominence} and have a geographic distance. A more general example for an om-space is the set of Airports together with the connecting plane routes relation and the  geodesic distances.%

The analysis of ordinal data is often based on a (numerical) metrification. Imposing metric structures on ordered sets\footnote{In the realm of machine learning a natural consequence when learning an ``embedding''.} is a widely studied~\cite{MONJARDET1981173}, yet still very active, field of research. Almost all mathematical disciplines contribute to this, e.g., algebra~\cite{communicabilityDistance}, combinatorics and geometry~\cite{linial1995geometry}, and analysis~\cite{analysisMetricGraph}. Furthermore, the task of (distance) metric learning in the realm of (weakly) supervised machine learning addresses a similar problem, e.g., based on the generalized \emph{Mahalanobis distance}.  Contemporary developments~\cite{distanceLearningSurvey} do often use deep learning settings. However, recent works claim that the reported achievements are rather marginal~\cite{musgrave2020metric}. We contributed to the task of finding meaningful metrics in ordinal data in two ways. First, we adapted the notions of dominance and prominence from the research field of orometrics to network science~\selfcite{schmidt2018prominence}. This process revealed a whole class of new metrics to study. Second, we have further developed the popular \emph{word2vec} method, which is based on a neural network model, for the special requirements of (concept) lattices~\groupcite{durrschnabel2019fca2vec}.

\randbemerkung{Euclidean Space}
A popular embedding structure for ordinal and metric data in the realm of machine learning is the Euclidean space, especially represented as $\mathbb{R}^n$. Depending on the properties of the relation (symmetric, transitive, reflexive, etc.) and the kind of embedding, e.g., \emph{distance}, \emph{similarity}, or \emph{transitional}, different bounds for $n$~\cite{pmlr-v117-bhattacharjee20a} are required for distortion free, i.e., isometric, embeddings. Often $n$ equals $|P|$, which may be computationally intractable. Moreover, it is a well known fact~\cite{MAEHARA20132848} that many finite metric spaces are not distortion-free embeddable into $\mathbb{R}^n$, for any $n\in\mathbb{N}$. A simple example for this is a 4-element ring with the graph distance as a metric. Therefore, low distortion embeddings~\cite{ABRAHAM20113026} and different means of measuring distortion~\cite{conf/nips/VankadaraL18} are widely studied. Hence, the investigation of distortions for embedding finite om-spaces into $\mathbb{R}^n$, which takes into account metric and order relational characteristics, is at hand, highly ambitious, and up to our knowledge, not conducted. We have already found initial evidence in a special case that the distortion can be extensive~\cite{DBLP:conf/iccs/HanikaH21}.

\randbemerkung{Gromov-Hausdorff Distance}
Understanding different metric embeddings of the same ordinal data is possible by comparing the resulting metric spaces. Amongst others, the Gromov-Hausdorff (GH) distance (by D.~Edwards~\cite{edwards1975structure,tuzhilin2016invented}, M.~Gromov~\cite{gromov1981groups}) enables such a comparison naturally. However, since GH is computationally intractable~\cite{memoli2007} one might resort to computable lower bound methods, e.g., the modified GH distance~\cite{memoli2012some}, which is time polynomial.

To further mitigate the computational demands in the last paragraphs, we can refer the reader to various works we carried out, e.g., the identification of important features in relational data~\selfcite{conf/iccs/HanikaKS19} based on entropy maximization and structure preservation, dataset size reduction using cores in formal contexts~\groupcite{hanika2020knowledge}, and the identification of clones~\selfcite{conf/ismis/DoerfelHS18}, \ie concept lattice preserving permutations. All works are  applicable to om-spaces, as every order relation gives rise to a (unique) concept lattice; and every metric on $P$ can be lifted to the power set.

To the best of our knowledge, studies purely concerned with the interplay of order relations and metrics with respect to data analysis are rare. We can identify in the literature few results about fix points~\cite{zbMATH06583504,bhaskar2006fixed} and contractions~\cite{agarwal2008generalized}. Related topological results for ordered metric spaces are of less interest for our envisioned task,  as the considered om-spaces are almost always finite, which in turn implies that the induced topological space is discrete.

\randbemerkung{Intrinsic Dimension}
Coming full circle with our motivation from mm-spaces, we propose to consider different results on valuation functions in ordered sets. This topic is already studied for order relations and lattices~\cite{MONJARDET1981173,KWUIDA2011990}. We already contributed to this by transferring ideas from orometrics to bounded metric spaces endowed with binary relations~\selfcite{stubbemann2020orometric,stubbemann2023montblanc}. Even more related to the realm of mm-spaces is our work on an intrinsic dimension of geometric datasets~\selfcite{hanika2018intrinsic_geometric,hanika23}. Although the proposed dimension function relies on mm-spaces, it is fully capable for the dimension analysis of om-spaces.

Finally, comparing real-world data with  randomly generated data often reveals new structural insights and provides benchmarks. %
Yet, uniform sampling of ordered sets is a difficult problem, at least for reasonable set sizes~\cite{KOZIEL2020294}. Our previous work allows to some extent for randomly generating binary relations~\groupcite{conf/iccs/FeldeH19} and null-model generation for (order) relations~\selfcite{felde2020null}. To which extent these methods can be useful for the study of om-spaces is an open question.

\subsection{Promising Research Questions}
Central to this research is the theoretical and experimental study of the triple $(P,R,d)$, as introduced in the last section as \fett{ordered metric space (om-space)}. One should distinguish between originally metric data (and therefore spaces)  and those where the metric was obtained through an external (\eg machine learning) method. Special interpretations of this structure are, for example, directed acyclic graph metric spaces (if the metric values are additive along paths), or weighted directed acyclic graphs (when considering metric values for connected pairs, i.e., relational pairs, only). One should approach the goal for a structure theory of om-spaces for ordinal data science from different angles, most importantly graph theory, order theory, and the analysis of metric spaces.

Any new result will surely build up on modelings concerned with $R$ being a tree or forest relation and $d$ a graph metric on it. The guiding methodology should be to examine the extensive apparatus of order theory with respect to the om-space. In particular, it is necessary to develop new concepts and methods to assess algorithmically the compatibility of $R$ and $d$. A special focus should be set on understanding if and how different kinds of dimensionality (e.g., order dimension, intrinsic dimension) do inflict the compatibility of $R$ and $d$ as well as a possible later embedding of $(P,R,d)$ into Euclidean space.

\subsubsection{Structure Theory of om-Space  Datasets}
\randbemerkung{Distortion}

To explore the distortion of order relations on sets with respect to indigenous and externally imposed metric functions is the focus of this research task. A first attempt to assess the distortion of an order relation in this sense can be based on the distortion of a map $\varphi$ between two metric spaces $(X,d_X)$ and $(Y,d_Y)$. In the realm of machine learning numerous distortion functions are studied~\cite{conf/nips/VankadaraL18}. A promising start may be with  $\mathop{dis}(\varphi) := \sup_{x,\hat x\in X} |d_X(x,\hat x)-d_Y(\varphi(x),\varphi(\hat x))|$.
To study the \emph{distortion introduced by relation $R$ in om-space $(P,R,d)$}, one could then study the canonical (by means of the Galois connection) map  $\varphi:P\to 2^P$ defined by $x\mapsto \{y\in P\mid (x,y)\in R\}$ from metric space $(P,d)$ to $(2^P, d_H)$ where $d_H$ is the Hausdorff metric derived from $d$.
This setting allows for analytical tools to assess the relational distortion (as well as expansion, extraction, etc.) of ordered metric spaces for real world datasets.
If $R$ is a purely reflexive relation, i.e., the simplest order relation on $P$, there will be no distortion, i.e., $\mathop{dis}(\varphi)=0$. Yet, for arbitrary relations in real-world datasets the change of distortion is unknown. 

\begin{figure}[t]
  \begin{center}
  \begin{minipage}[t]{0.4\textwidth}
    \begin{cxt}
      \small
      \cxtName{$\context$}
      \atr{Aeroflot}
      \atr{Air France}
      \atr{Lufthansa}
      \atr{Scandinavian}
      \atr{British Airways}
      \atr{Austrian A.}
      \atr{Alitalia}
      \obj{xxxxxxx}{Hamburg}
      \obj{.xx.xxx}{Lisbon}
      \obj{xxxxx.x}{Madrid}
      \obj{xx.x..x}{Moscow}
      \obj{.xx.x.x}{Toulouse}
      \obj{xxxxxxx}{Budapest}
      \obj{xxx....}{Dresden}
      \obj{xxxxxxx}{London}
      \obj{xxx.x.x}{Marseille}
      \obj{xxxxxxx}{Rom}
      \obj{.xxxxxx}{Palma D.M.}
      \obj{..x..x.}{Leipzig/Halle}
    \end{cxt}
  \end{minipage}
  \hfill
  \begin{minipage}{0.54\textwidth}
    \trimbox{0cm 2.35cm 0.25cm 0cm}{\scalebox{0.88}{\colorlet{mivertexcolor}{blue}
\colorlet{jivertexcolor}{red}
\colorlet{vertexcolor}{mivertexcolor!50}
\colorlet{bordercolor}{black!80}
\colorlet{linecolor}{gray}
\tikzset{vertexbase/.style 2 args={semithick, shape=circle, inner sep=2pt, outer sep=0pt, draw=bordercolor},%
  vertex/.style 2 args={vertexbase={#1}{}, fill=vertexcolor!45},%
  mivertex/.style 2 args={vertexbase={#1}{}, fill=mivertexcolor!45},%
  jivertex/.style 2 args={vertexbase={#1}{}, fill=jivertexcolor!45},%
  divertex/.style 2 args={vertexbase={#1}{}, top color=mivertexcolor!45, bottom color=jivertexcolor!45},%
  conn/.style={-, thick, color=linecolor}%
}
\begin{tikzpicture}[scale=0.3]
  \begin{scope} %
    \begin{scope} %
      \foreach \nodename/\nodetype/\param/\xpos/\ypos in {%
        0/vertex//0.27679769894522366/3.316107382550335,
        1/jivertex//-2.0/6.0,
        2/jivertex//-2.0/10.0,
        3/jivertex//6.659539789069875/11.169031639501437,
        4/jivertex//-8.0/14.0,
        5/jivertex//-2.0/14.0,
        6/vertex//4.328763183125503/14.001821668264622,
        7/jivertex//8.703451581974967/14.001821668264622,
        8/vertex//-8.0/18.0,
        9/vertex//4.866634707574214/19.021955896452543,
        10/divertex//9.456471716203133/21.675455417066157,
        11/mivertex//-8.0/22.0,
        12/mivertex//-2.0/22.0,
        13/vertex//5.404506232022936/22.715340364333656,
        14/mivertex//-1.659539789070017/26.62387344199425,
        15/mivertex//5.512080536912709/26.65973154362416,
        16/mivertex//-1.3726749760307158/30.6758389261745,
        17/vertex//-1.2651006711409778/35.05052732502397
      } \node[\nodetype={\param}{}] (\nodename) at (\xpos, \ypos) {};
    \end{scope}
    \begin{scope} %
      \path (4) edge[conn] (12);
      \path (4) edge[conn] (8);
      \path (16) edge[conn] (17);
      \path (0) edge[conn] (1);
      \path (0) edge[conn] (3);
      \path (5) edge[conn] (13);
      \path (5) edge[conn] (11);
      \path (6) edge[conn] (12);
      \path (6) edge[conn] (9);
      \path (2) edge[conn] (5);
      \path (2) edge[conn] (8);
      \path (2) edge[conn] (9);
      \path (7) edge[conn] (10);
      \path (7) edge[conn] (9);
      \path (10) edge[conn] (15);
      \path (1) edge[conn] (4);
      \path (1) edge[conn] (6);
      \path (1) edge[conn] (2);
      \path (12) edge[conn] (14);
      \path (13) edge[conn] (16);
      \path (13) edge[conn] (15);
      \path (3) edge[conn] (6);
      \path (3) edge[conn] (7);
      \path (8) edge[conn] (11);
      \path (8) edge[conn] (14);
      \path (11) edge[conn] (16);
      \path (15) edge[conn] (17);
      \path (14) edge[conn] (16);
      \path (9) edge[conn] (13);
      \path (9) edge[conn] (14);
    \end{scope}
    \begin{scope} %
      \foreach \nodename/\labelpos/\labelopts/\labelcontent in {%
        9/above//{\emph{British A.}},
        10/above right//{\emph{Austrian A.}},
        11/above left//{\emph{Aeroflot}},
        12/above//{\emph{Scand.}},
        14/above//{\emph{Alitalia}},
        15/above right//{\emph{Lufthansa}},
        16/above left//{\emph{Air France}}
      } \coordinate[label={[\labelopts,yshift=-0.3cm]\labelpos:{\labelcontent}}](c) at (\nodename);
    \end{scope}
     \begin{scope} %
      \foreach \nodename/\labelpos/\labelopts/\labelcontent in {%
        1/below//{Madrid},
        2/below//{Marseille},
        3/below right//{Palma D.M.},
        4/below//{Moscow},
        5/below//{Dresden},
        7/below//{Lisbon},
        9/below//{Toulouse},
        10/below right//{Leipzig/Halle}
      } \coordinate[label={[\labelopts,yshift=0.3cm]\labelpos:{\labelcontent}}](c) at (\nodename);
      \coordinate[label={[yshift=1.5cm] below right:{Hamburg, Budapest London, Rom}}] (c) at (0);
    \end{scope}
  \end{scope}
\end{tikzpicture}}}
  \end{minipage}
  \end{center}
  
  \caption{\textbf{Left:} Simplified formal context from real-world data about cities (objects) that are serviced by airlines (attributes). \textbf{Right:} Corresponding concept lattice.}
  \label{fig:airlinescontext}
\end{figure}
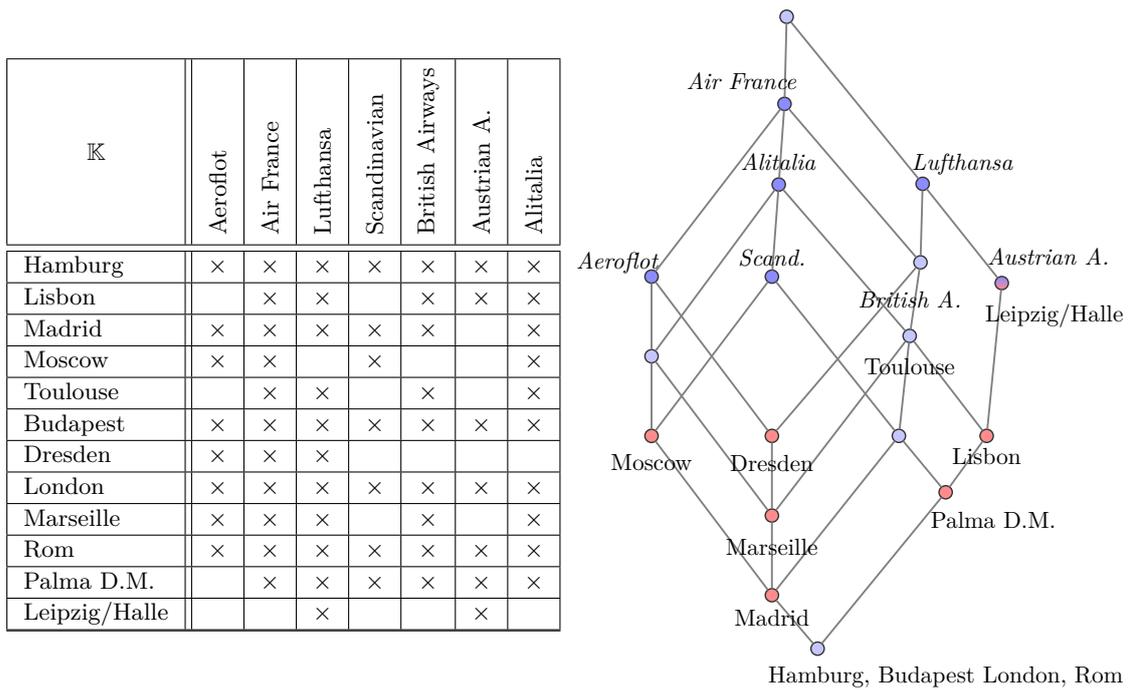

\begin{example}
  \label{ex:omspace}
  The data shown in Figure~\ref{fig:airlinescontext} and Table~\ref{tab:distances-airports} can be comprehended as an om-space. The elements of $P$ are the cities, the map $d$ assigns any two cities their geodesic distance, and any two cities are in relation $R$ iff there is an airline servicing both cities. \qed
\end{example}

\begin{table}[t]
  \caption{Geodesic distances between cities (in nautical miles [nmi]) that
    are the objects of the formal context in
    Figure~\ref{fig:airlinescontext}.}
  \centering
  {\scriptsize
\begin{tabular}{lrrrrrrrrrrrr}
\toprule
{} & \rotatebox{90}{Hamburg} & \rotatebox{90}{Lisbon} & \rotatebox{90}{Madrid} & \rotatebox{90}{Moscow} & \rotatebox{90}{Toulouse} & \rotatebox{90}{Budapest} & \rotatebox{90}{Dresden} & \rotatebox{90}{London} & \rotatebox{90}{Marseille} & \rotatebox{90}{Rom} & \rotatebox{90}{Palma D.M.} & \rotatebox{90}{Leipzig/Halle} \\
\midrule
Hamburg       &                       0 &                   2198 &                   1781 &                   1765 &                     1278 &                      950 &                     378 &                    745 &                      1186 &                1326 &                       1658 &                           289 \\
Lisbon        &                    2198 &                      0 &                    513 &                   3892 &                     1029 &                     2480 &                    2250 &                   1564 &                      1307 &                1839 &                       1027 &                          2174 \\
Madrid        &                    1781 &                    513 &                      0 &                   3420 &                      537 &                     1975 &                    1784 &                   1246 &                       797 &                1330 &                        546 &                          1718 \\
Moscow        &                    1765 &                   3892 &                   3420 &                      0 &                     2883 &                     1563 &                    1643 &                   2508 &                      2666 &                2399 &                       3119 &                          1722 \\
Toulouse      &                    1278 &                   1029 &                    537 &                   2883 &                        0 &                     1453 &                    1249 &                    883 &                       312 &                 911 &                        468 &                          1188 \\
Budapest      &                     950 &                   2480 &                   1975 &                   1563 &                     1453 &                        0 &                     573 &                   1490 &                      1180 &                 836 &                       1590 &                           674 \\
Dresden       &                     378 &                   2250 &                   1784 &                   1643 &                     1249 &                      573 &                       0 &                    988 &                      1070 &                1044 &                       1546 &                           111 \\
London        &                     745 &                   1564 &                   1246 &                   2508 &                      883 &                     1490 &                     988 &                      0 &                       989 &                1444 &                       1348 &                           879 \\
Marseille     &                    1186 &                   1307 &                    797 &                   2666 &                      312 &                     1180 &                    1070 &                    989 &                         0 &                 602 &                        479 &                          1032 \\
Rom           &                    1326 &                   1839 &                   1330 &                   2399 &                      911 &                      836 &                    1044 &                   1444 &                       602 &                   0 &                        839 &                          1070 \\
Palma D.M.    &                    1658 &                   1027 &                    546 &                   3119 &                      468 &                     1590 &                    1546 &                   1348 &                       479 &                 839 &                          0 &                          1511 \\
Leipzig/Halle &                     289 &                   2174 &                   1718 &                   1722 &                     1188 &                      674 &                     111 &                    879 &                      1032 &                1070 &                       1511 &                             0 \\
\bottomrule
\end{tabular}}

  \label{tab:distances-airports}
\end{table}

Applying this new measure on real-world data would require non-incremental research to develop efficient algorithms to compute them, as computational demands increase drastically with increasing $|P|$. One might face this highly ambitious challenge by drawing from results on closed sets, which arise naturally from the closure operator generated by $R$. Hence, one may employ fast algorithms for the generation of closed sets (or their generators), which will facilitate the computation of the Hausdorff distances for large sets $P$. Starting from this promising modeling, the next step would then be to  test, examine and develop further distortion measures in the realm of om-spaces.

\subsubsection{Formal Context Mediated Metrics and Orders by Valuation}
In a more general setting one might also want to study arbitrary binary relations, as often investigated in the form of a formal context $(G,M,I)$. In this regard, the following question is directly obvious. How does the binary incidence relation $I\subseteq G\times M$, and therefore the inherent lattice structure, mediate a metric from one set to another? In detail, given the incidence $I$ between sets $G$ and $M$, where $G$ is equipped with a metric $d$, one can induce a metric on $M$ using the afore mentioned Hausdorff metric on $(G,d)$ and the map $m\mapsto \{g\in G\mid (g,m)\in I\}$. What properties are expected of the resulting metric space $(M,d_{M})$? What alterations of $I$, i.e., operations based on adding or removing relational pairs, do change the properties of $(M,d_M)$? Based on this one might  also want to investigate the special case in which both sets $G$ and $M$ are equipped with an indigenous metric. Then, the natural question arises: how compatible are these indigenous metrics with the mutually induced ones mediated by the incidence $I$?
For this, the computation of distance between metric spaces is essential, for which it might be necessary to employ the modified GH distance~\cite{memoli2012some}.

\begin{example}
  It is clear that in Example~\ref{ex:omspace} only the object set $G$
  is equipped with a distance function (actually a metric) $d_{G}$. In order to obtain a distance function between the airlines, i.e. the attributes, the most obvious way is to calculate a Hamming distance between them. This distance, however, would not consider the existing data for $G$. Based on the proposition above, we can infer a distance function that incorporates $d_{G}$. For example, $d_{M}(\text{Scandinavian},\text{Austrian})$ would be mapped to the Hausdorff distance $d_{H}(M_{1},M_{2})$ where
  \begin{align*}
    M_{1}&=\{\text{Hamburg, Madrid, Moscow, Budapest, London, Rom, Palma D.M.}\}\\
    M_{2}&=\{\text{Hamburg, Lisbon, Budapest, London, Rom, Palma D.M., Leipzig/Halle}\}.
  \end{align*}
  Hence, one has to compute
  \[\max\left\{\sup_{x\in M_{1}}d(x,M_{2}),\sup_{y\in M_{2}}d(M_{1},y)\right\},\]
  where $d(u,V)\coloneqq \inf_{v\in V}d(u,v)$, i.e., the distance from
  $u$ to $V$. In our example the computation of the Hausdorff distance
  results in $\max\{836, 1563\}=1563$\,nmi.\qed
\end{example}

\randbemerkung{Valuation Functions}
Closely related to the setting above is the application of valuation functions to generate (linear) order relations. An example is to rank objects from $G$ based on sizes of their shared related subsets from $M$, i.e., every $g\in G$ is valued by $|\{m\in M\mid (g,m)\in I\}|$. Since valuation functions almost always map into a numerical scale, e.g., the real line, this results in a metric and in turn into a linear order. Dual to the question in the first paragraph one can investigate how a valuation map distorts an indigenous ordinal property of the set, for example extracted through the notion of concept lattices. As a start, one may count how many wrongly ordered relation tuples are present. Furthermore, it might be fruitful to  investigate to which  extent the order created by the valuation differs from an arbitrarily generated linear extension~\cite{bubley1999faster}. Finally, one might  re-evaluate our preliminary work in this area~\selfcite{schmidt2018prominence,stubbemann2020orometric,stubbemann2023montblanc}. This can shed light on the question whether the valuation functions we transferred from orometry are superior when dealing with indigenous ordinal properties.

\subsubsection{Intrinsic Dimension of om-Spaces and Randomness}
\randbemerkung{Intrinsic Dimension}
There are numerous notions for measuring the complexity of a dataset. For ordinal data the order dimension is a natural candidate, as described in Section~\ref{sec:ordered-sets}. Another approach is the notion for \emph{intrinsic dimension}. Put simply, this value reflects the number of dimensions required to accurately represent the data. For incidence structures, such as formal contexts, we proposed in our previous work ``Intrinsic Dimension of Geometric Data Sets''~\selfcite{hanika2018intrinsic_geometric} a novel intrinsic dimension function that incorporates measure- and metric properties. This function is also applicable to ordinal data. This is in particular true, if the ordinal data is scaled via the \emph{general ordinal scale}.

Yet, it is unknown if the intrinsic dimension does relate to the order dimension in any way. Still, it is to be expected that the intrinsic dimension can be used to capture ordinal complexity to some extent. To substantiate the hypothesis, an extensive experimental evaluation is needed and should be conducted, using real-world as well as randomly generated datasets. For the latter we might refer the reader to our work on random relational data generation~\groupcite{conf/iccs/FeldeH19}. Moreover, one can compare our results to recent notions of estimating dimensionality in the absence of distance functions~\cite{kleindessner2015dimensionality}. Building up on this evaluation, and even more progressive, is the further development of an extension of the intrinsic dimension to om-spaces. This extension can open up a new research avenue for dataset dimensionality --- however, it is dependent on highly ambitious theoretical and algorithmic results from all paragraphs above.

\section{Algebraic constructions/decompositions for complexity reduction}
\label{sec-algebraic-constructions}

For algebraic structures such as lattices, there exist many ways of reducing or decomposing them to smaller parts, for instance factorizations or sub-direct decompositions. However, these constructions are very sensitive to small perturbations in the data; and approximations and heuristics in the style of data mining are not (yet) considered in algebraic research.\footnote{There are other approximations in universal algebra, but of a different nature, which is not relevant here: they deal with approximating infinite algebras \cite{meinke1993universal} and serve, \eg for defining the semantics of programming languages.}
The challenge is thus to \fett{equip the algebraic constructions with approximations and heuristics} s.\,t. they become \fett{suitable for data analysis.}

We propose to develop methods for clustering, aggregating and factorizing ordered sets and lattices. For ordered sets this may include order-based modifications of clustering algorithms such as $k$-modes, hierarchical agglomerative clustering and density-based clustering; and one might explore which signals for clustering can be obtained from their Dedekind-MacNeille completions. For lattices, we suggest exploiting constructions of universal algebra and lattice theory (such as factorizations, tolerance relations, atlas decompositions etc.) \cite{davey2002introduction} \selfcite{stumme98free}. These methods, however, have not been developed with real-world data with noise or disturbances in mind --- up to now there are no notions such as ``the lattice is decomposable to 95\,\%'', and no algorithms dealing with them.

\subsection{State of the Art and Open Questions}

\noindent
\randbemerkung{Universal algebra as theoretical toolkit for data analysis}
In universal algebra \cite{birkhoff1946universal,gratzer1968universal,burris1981course} it is a well-known fact that every homomorphic image and every subalgebra of a lattice or of a finite complete lattice and every direct product thereof is a lattice again~\cite{birkhoff1935structure}.\footnote{The same holds for all classes of algebras that are defined solely by equations, \eg for groups and rings.} This means that universal algebra provides us already with a rich toolkit for decomposing and aggregating lattices. In particular, every congruence relation of the lattice (\ie an equivalence relation that is compatible with meet and join) provides a clustering of the lattice that additionally respects the lattice operations. The interpretation as clustering works much better for lattices than for other types of algebras (\eg groups), as each congruence class is a convex set (``cluster'') and is therefore not shattered throughout the lattice. The corresponding factor lattice can be understood as an aggregate of the original lattice. Several approaches exist that weaken the requirement of a congruence relation. Tolerance relations \cite{Wille85a}, for example, allow for overlapping clusters, while our newly introduced interval relations \cite{koyda2023factorizing} produce smaller clusters (on the cost of not respecting all meets and joins any more).
Another way of reducing the complexity of a lattice is by exploiting the fact that every lattice can be decomposed into smaller, irreducible factors.

\randbemerkung{Lattice composition and decompositions}
Many of these algebraic operations were linked to datasets (in the form of a formal context) within the field of FCA~\cite{Ganter88}; in particular for subdirect decompositions \cite{Reuter86,Reuter87}, for direct products of convex-ordinal scales \cite{Strahringer94}, for substitution decomposition \cite{LukschWille87,Stephan91b} and substitution product \cite{Wille87a,Stephan91a}, as well as for tensorial decomposition \cite{Wille85b} and tensor products \cite{Wille91c}.

\begin{figure}
	\includegraphics[width=\textwidth]{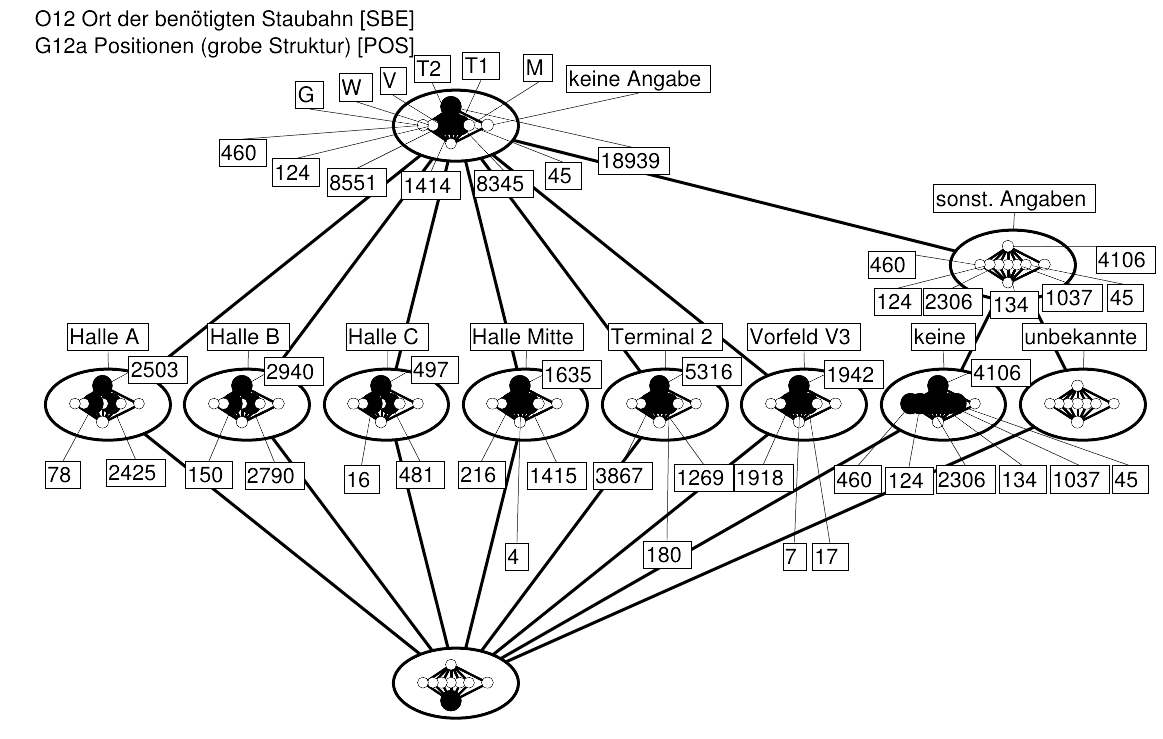}
	\caption{Subdirect product of the scales \emph{position of baggage conveyor} and \emph{position of aircraft} at Frankfurt Airport}
	\label{fig-staubahn}
	\end{figure}

\begin{example}
Figure~\ref{fig-staubahn} shows the subdirect product of two conceptual scales for a conceptual information system about flight movements at Frankfurt Airport. With this composition of scales, one can analyze the distribution of the flight movements over different dimensions. 
In the figure, the outer scale indicates the location of the baggage conveyor assigned to an aircraft. For instance, we can see at the left of the diagram that 2503 times a baggage conveyor in Halle A was assigned to an aircraft. The inner scale shows the position of the aircraft. Of the 2503 aircraft mentioned before, 78 were positioned on the apron (V\,=\,Vorfeld) and 2425 at Terminal 1 (T1). Both assignments are reasonable because Halle A is part of Terminal 1. However, the system also helps us to discover apparent mismatches. For instance, we find that there are 180 aircraft that had a baggage conveyor assigned at Terminal 2 but which were positioned at Terminal 1 (T1). Focussing on these 180 flight movements with further scales would support us to discover reasons for this apparent mismatch. 
	\qed\end{example}
\randbemerkung{Ordinal Clustering and Factorization}
Complete congruence relations of concept lattices \cite{ReuterWille87,KnechtWille90} are a good starting point for clustering and aggregating. Complete tolerance relations and atlas decompositions \cite{Wille85b} have been developed explicitly for allowing large lattices to be visualized in smaller parts just like several map pages in a road atlas.
\randbemerkung{Distributivity}
Decompositions and factorizations are more straightforward the more structure the lattice shows. A particular strong structure is distributivity $\left(\; x\wedge(y \vee z) = (x\wedge y) \vee(x\wedge z) \mbox{ and } x\vee(y\wedge z) = (x\vee y) \wedge (x\vee z)\;\right)$ \cite{birkhoff1935structure,Wille85c,davey2002introduction}. In this case, the lattice can be subdirectly decomposed into linear orders (which is beneficial for ordinal factor analysis as is discussed in Section~\ref{sec-ordinal-factor-analysis}).

\randbemerkung{Algebraic methods are sensitive to impurity}
However, many of these constructions have never been used for larger data analysis tasks. This is due to the fact that lattices resulting from real-world data often do not precisely fulfil the structural conditions required for applying a decomposition or a factorization. In such a case, a lattice is considered as irreducible; there is no such notion as ``95\,\% distributive'' or ``reducible with an error of $x$''. Techniques for dealing with partly imperfect data (as they exist in numerical settings, as for instance soft margins for support vector machines) are not in the spirit of universal algebra, and have thus not been developed so far. The same holds for distributivity, which is rarely observed in the whole concept lattice of real-world data. Nevertheless, empirical evidence shows that many lattices contain `a large distributive part' --- a still to-be-defined concept.

The main scientific challenge in this field is thus to \fett{develop methods for applying algebraic constructions to lattices even if the necessary structural conditions are not satisfied.} This will include the definition of measures for the degree of satisfaction of these conditions, the provision of preprocessing methods for `cleaning' the data appropriately, the establishment of a theory for compositions, decompositions and aggregations with structural impurity, and the development of algorithms and heuristics for their computation on large datasets.

\subsection{Promising Research Questions}
We propose three specific research questions in this area. The first two consider clustering in ordered sets and lattices, resp., while the third one focuses on decompositions of lattices. Considering the required research approaches, it is the latter two questions that resemble more, as both can draw on the richer structure of lattices. In those two, the challenge is how to either `structurally clean up' the data before applying the algebraic operations, or how to turn the operations more `fault-tolerant'.

\subsubsection{Clustering in Ordered Sets}
\begin{figure}
	\includegraphics[width=\textwidth]{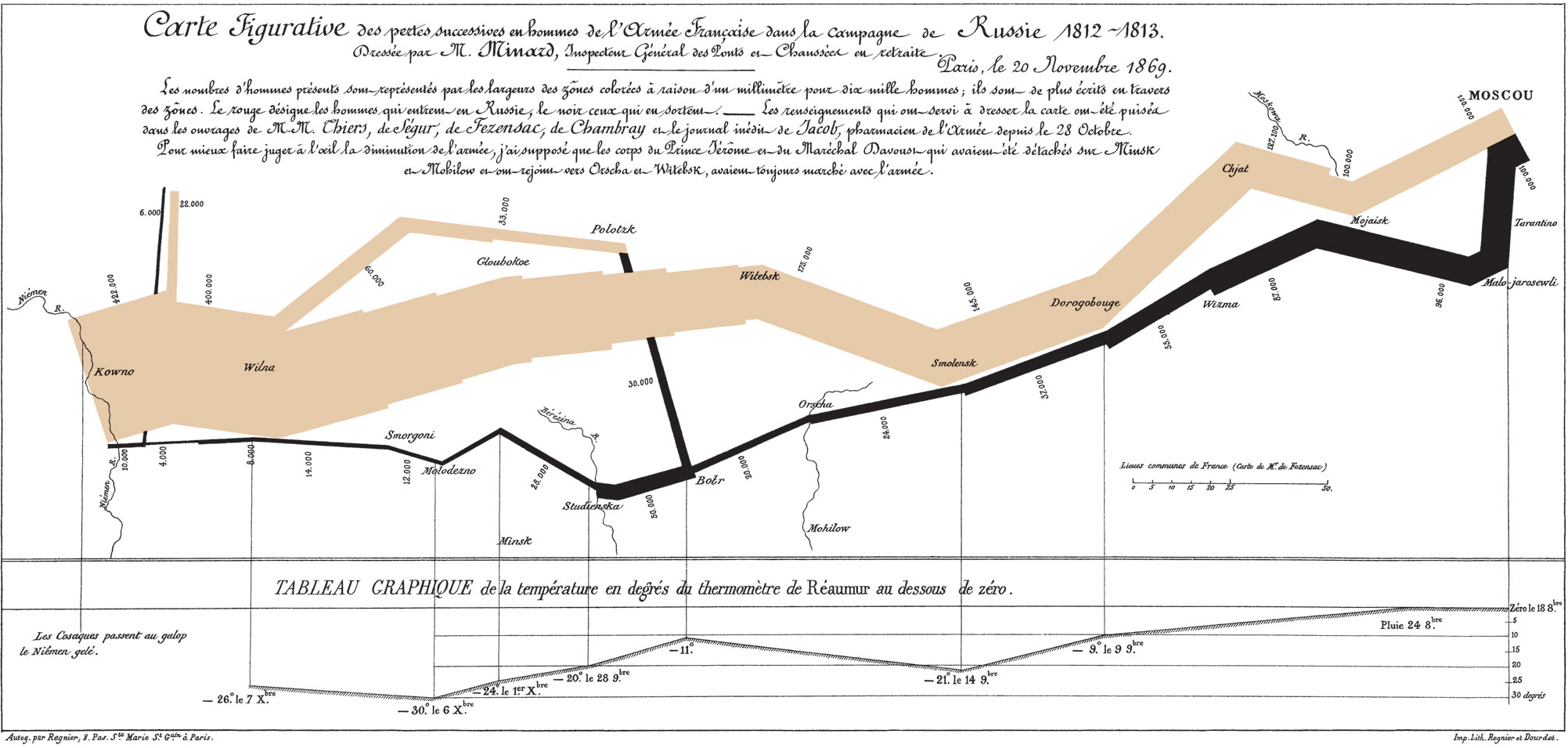}
		\caption{Minard's map of Napoleon's campaign in Russia as inspiration for a possible density-based clustering algorithm of an ordered set.}
		\label{fig-minard}
	\end{figure}

There are (at least) two straightforward approaches for clustering algorithms for ordered sets, the first being the adaptation of classical clustering techniques to ordered sets, and the second being the direct exploitation of order-theoretic constructs. For the adaption approach, we suggest to start with exploring modifications of standard clustering algorithms that rely on the comparability of pairs of objects, as opposed to numerical distances. These include for instance $k$-modes \cite{593469} and hierarchical agglomerative clustering \cite{10.2307/2282967} (where `hierarchical' does refer to the set inclusion hierarchy of the clusters). We also consider it extremely promising to develop ordinal versions of modularity-based clustering of networks \cite{Blondel2008FastUO}, and --- inspired by Ch. Minard's seminal map of Napoleon's campaign in Russia (Figure~\ref{fig-minard}, \cite{minard1869carte})  --- ordinal versions of density-based clustering~\cite{ester1996densitybased}.

\randbemerkung{Ordinal Clustering}
In the second approach, one might investigate how to exploit order-theoretic concepts (such as common upper and lower bounds, order filters and ideals, and intervals) for new kinds of ordinal clustering. Particularly promising is to consider the new nodes in a Dedekind-MacNeille completion as cluster nuclei. This is due to the observation that, in the \texttt{ancestor$\_$of} hierarchy of a human genealogy, a couple with two or more children always generates a new node in the Dedekind-MacNeille completion, which may be interpreted as a node representing the family.

\subsubsection{Clustering and Factorizations in Lattices} 
If an ordered set is a lattice (including the case when we enforce this by the Dedekind-MacNeille completion), one can benefit from more structure. As stated above, in theory congruence relations and tolerance relations already provide clusterings that are even compatible with the lattice operations. However, they may not always be applicable. Again, there are (at least) two promising approaches to remediate this:
The first is to assess the proportion of a lattice that has to be modified such that a congruence relation or a tolerance relation with pre-defined requirements exists; and derive methods to efficiently identify such modifications.
The second is to weaken the notions of congruence relations and tolerance relations, so that they may respect only most of the operations. The questions here are again how to define them, how to compute (or approximate) them efficiently, and how to interpret the results. This may result in a new notion of `operator-aware modularity'. An expected problem case in modularity based graph clustering is a cluster that is a fully connected subgraph. In lattices, this corresponds to (convex) Boolean sublattices. This correspondence could give rise to a theory and algorithms for their compression in a cluster. 

\randbemerkung{Closure Systems}
Another line of research one may draw from is employing other representation modes of the lattice: One could for instance study if there are meaningful concepts for clustering sets of implications, and how they translate back again to the clustering problem for lattices. In particular in the case of a large dataset one could also resort to probabilistic methods to derive the propositional logic and dependencies~\cite{Studeny}. One might also employ results from preliminary work on probably approximately correct learning~\groupcite{conf/icfca/BorchmannHO17,BORCHMANN202030} of those dependencies.

\subsubsection{Decompositions of lattices}
Concerning the complexity reduction of lattices by decompositions, one can follow the same research methodology as for clusterings and factorizations: There exists a large body of algebraic decomposition constructions for lattices, such as direct and sub-direct decompositions, or tensor decompositions \cite{ganter99formal}, but these do not admit for minor structural disturbances. Hence, a natural task would be to adapt them to more robust versions, either by suitable preprocessing or by modifying the decomposition mechanism directly.

\randbemerkung{``Almost distributive''}
A promising observation with respect to preprocessing is that many real-world lattices are ``to a large extent'' distributive at their top \cite{DBLP:conf/iccs/StrahringerWW01}, and that distributive lattices allow for particularly simple decompositions (\eg subdirect decomposition into linear orders). Hence, the question for how to `repair' a lattice s.\,t.\ it becomes distributive is meaningful.
In the case of ``almost distributive'' lattices, one could resort to the study of how (potentially non-distributive) ordered sets can be freely completed (in the sense of universal algebra) to lattices~\selfcite{stumme98free}. A potential means to identify the non-distributive part of a lattice might then be to establish a way to reverse this construction and develop efficient computation methods.

Additionally, one might want to explore --- as described above --- how other equivalent representations (in particular sets of implications) can be decomposed, and how this can be translated into meaningful decompositions of lattices.

\section{Ordinal Factor Analysis}
\label{sec-ordinal-factor-analysis}

Factor analysis (in its traditional sense) is a method of multi-variate statistics to reduce data consisting of observations in different manifest variables to a lower-dimensional space spanned by `latent variables' with as little loss as possible. As the variables are considered to range in the real numbers, we will call this task numerical factor analysis in the sequel. A variety of approaches exists, the most prominent being principal component analysis. However, all these approaches require the data to be at least on interval scale level. Ordinal factor analysis, on the other hand, has been developed with the same intention as its numerical counterpart, but focussing on data of ordinal scale type.

Based on this foundation, we are looking forward to a
\fett{comprehensive theory and algorithms for an ordinal version of factor analysis}, that will keep track of all operations and thus avoid the problem of numerical factor analysis where distances may be distorted.
For ordinal data the factors are known to be ordered sets. One might continue research in this direction by focussing on linear ordered sets as factors, as they provide an intuitive representation, in particular when their number is low. This constitutes an instance of dimension reduction and is computationally hard for ordinal data. Hence, one may have to resort to approximate methods for real-world data applications. In a second step, the theoretical foundation of ordinal factor analysis might then be adapted to non-linear factors.
Our ambition is to establish ordinal factor analysis as a tool for ordinal data analysis, that  will provide \fett{a more intuitive representation of complex ordinal data}.

\subsection{State of the Art and Open Questions}
In multivariate statistics empirical observations are used to draw conclusions about underlying independent random variables.
This was first developed by Spearman~\cite{spearman1904proof} in 1904 for evaluating intelligence tests, tracing them back to a single ``general factor'', and was generalized to multiple variables in 1919 by Garnett~\cite{Garnett.1919}.
Building on this notion, nowadays two variants of factor analysis are employed.
On one hand, \emph{confirmatory factor analysis}~\cite{Harrington.2009} is used in social research to test whether the data fits to hypothesized measurement models.
On the other hand, \emph{explorative factor analysis}~\cite{Norris.2010} is used to discover hidden structures in underlying data without the numbers of factors being known beforehand.
This explorative factor analysis is closely related and often confused with principal component analysis as noted in \cite{Joliffe.1992}, even though both techniques differ in execution as well as in their aim.
In the area of recommender systems, factor analysis came to prominence through the `Netflix Prize' that was held between 2006 and 2009 where Funk~\cite{Funk.2006} showed how to use factor analysis techniques to predict users preferences of movies.
This approach was improved and refined in~\cite{Kumar.2014, He.2017}.
A similar but not equal way of doing factor analysis is called \emph{non-negative matrix factorization} and applied in cluster analysis of documents~\cite{Sra.2006} or in astronomy~\cite{Berne.2007}.
One problem of these approaches is that the resulting factors are difficult (or even impossible) to interpret for a data analyst.

\randbemerkung{Explainable Factor Analysis}
To avoid this problem, steps towards an explainable factor analysis have been made in the field of FCA.
This goes back to Kerpt and Sn{\'a}sel~\cite{Keprt.2004, Keprt.2006} and was further developed by Belohlavek and Vychodil~\cite{Belohlavek.2007, Belohlavek.2010} under the name \emph{Boolean factor analysis}.
They compute a factorization of a Boolean matrix into two binary matrices, such that their Boolean matrix product results in the original matrix.

For such a factorization there is a corresponding factorizing family in the lattice, consisting of $k$ formal concepts, called Boolean factors.
Even though it is known to be \NPcomplete{} to decide whether a Boolean matrix has a factorization into $k$ Boolean factors, they introduce a set of algorithms for Boolean factor analysis which optimize $k$.
Building on these notions, \emph{ordinal factor analysis} was introduced by Ganter and Glodeanu in~\cite{Ganter.2012,Ganter.2013}, as a technique to meaningfully group Boolean factors, together with a visualization technique to depict ordinal factors in a biplot.

\begin{figure*}[t]
	\centering
	\hspace*{-2.8em}
	\scalebox{0.78}{
		\raisebox{10em}{
			\begin{cxt}
				\atr{USA-based}
				\atr{premium}
				\atr{ads}
				\atr{private messages}
				\atr{group messages}
				\atr{mobile first}
				\atr{stories}
				\atr{timeline}
				\obj{x.xxx.xx}{Facebook}
				\obj{x.xxxxxx}{Instagram}
				\obj{xxxx....}{Reddit}
				\obj{x.xxxxx.}{Snapchat}
				\obj{...xxx..}{Telegram}
				\obj{..xxxxxx}{TikTok}
				\obj{xxxxxx.x}{Twitter}
				\obj{..xxxxx.}{WeChat}
				\obj{x..xxxx.}{WhatsApp}
				\obj{xxx...x.}{YouTube}
		\end{cxt}}
		\includegraphics[height=20em]{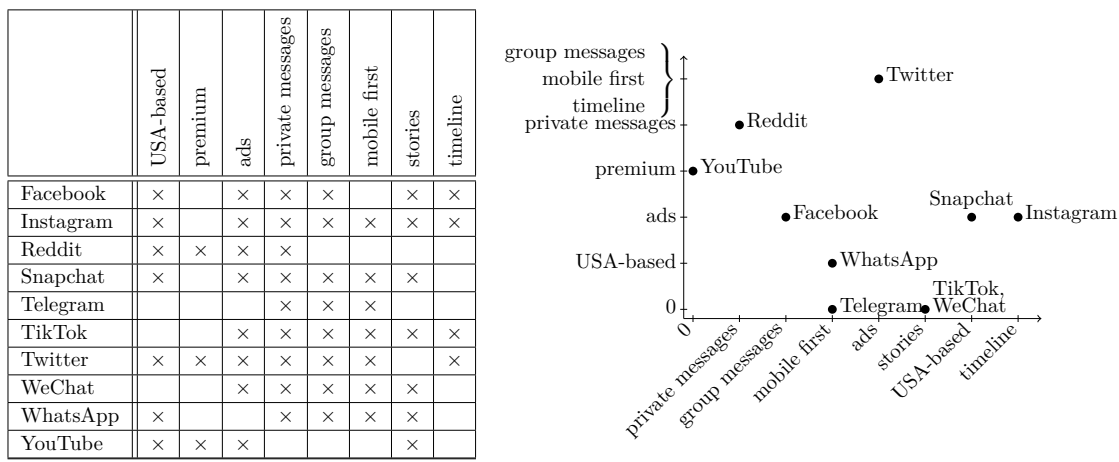}}
	\caption{A dataset on social networks together with a visualization of a factorization into its two largest ordinal factors. All incidences can be deduced from the projection except for (TikTok, timeline), (WhatsApp, stories), (Facebook, timeline), (YouTube, stories), (Facebook, stories). The ordinal projection does not contain false data.}
	\label{fig:2d}
\end{figure*}

\begin{example}
	Figure \ref{fig:2d} shows, on the left, a small dataset about prominent social networking platforms. On the right, a factorization of the dataset into its two largest ordinal data is shown. In the plot, an object contains all attributes that are listed below on the horizontal axis and all attributes that are listed left of it on the horizontal axis. The plot contains all information of the original dataset, except the four incidences listed in the caption of the figure. In order to cover these relations as well, the third factor would have to be included as well, leading to a 3D visualisation. \qed
\end{example}

The authors restrict their work to ordinal factors that form a chain in the concept lattice.
This \emph{linear factor analysis} is closely related to dimension theory of ordered sets~\cite{trotter2001combinatorics}, as in both cases a minimal number of chains is computed that covers an ordered set.
In~\cite{Glodeanu.2013_2} the authors show that linear factor analysis can be applied to data sets from real-world applications and in~\cite{Glodeanu.2013} the theory of linear factor is transferred to triadic concept analysis.
A modification of the visualization technique for the linear factors and the idea to extract two maximal linear factors that cover most of the data is proposed in~\cite{Ganter.2013}.
A way that tries to cover most of a concept lattice with two chains and can thus be seen as a variant of linear factor analysis is described in our work~\selfcite{durrschnabel2019drawing}.

While linear factor analysis seems to be a promising tool for a data analyst, it is still not yet fleshed out enough to be applied in practice.
This is mainly due to the fact of being computationally expensive, and thus algorithms have not yet been developed (except for the Boolean case).
Furthermore, the theoretical backbone of the theory is not yet broad enough to provide a data analyst with a robust toolkit of methods and variants.

\subsection{Promising Research Questions}

The overarching research question here is \fett{how to develop ordinal factor analysis to a mature (visual) data analysis tool, that can be applied using little manual effort}. In particular this calls for enriching the theoretical backbone and the development of algorithms to make ordinal factor analysis applicable in practice. The algorithms should be able to extract structure that is already encapsulated in data and to deal with incremental changes in the data.
Finally, a variant that also deals with non-linear factors (\eg trees) might be of interest.

\subsubsection{Computing Linear Factorizations in Two and More Dimensions}
A first step in this direction is our greedy algorithm~\cite{greedyfactor} that iteratively extracts the largest remaining ordinal factor from a formal context.
Furthermore, our second work~\cite{twofactor} builds on an idea proposed by Ganter~\cite{Ganter.2013} to cover a maximum subset of the dataset with a small set of linear factors.
This proposed method allows for the incorporation of heuristics to approximate a factorization into two dimensions, and examines the computational boundaries. %

\randbemerkung{Two Dimensional Linear Factor Analysis}
We assume two-dimensional linear factor analysis to perform better on smaller datasets, as large datasets tend to comprise more complex dependencies.
Thus, a second step may leap significantly beyond the state of the art by exploring how complex data can be represented with factor analysis in higher dimensions.
As the two-dimensional case is already known to be of high computational complexity, we do expect this to be an even more complicated problem, which will definitely require the development of heuristics.

As the aim of these factorizations is to improve the understanding of correlations in data sets for a data analyst, novel visualization techniques are required.
The practicability of resulting breakthrough approaches and algorithms of ordinal factor analysis in higher dimensions will have to be demonstrated.
As it hardly seems possible to visualize factors far beyond dimension three on a static two-dimensional document, interactive navigation methods will have to be developed as well.

\subsubsection{Background Orders and Evolving Data}

\randbemerkung{Background Orders}
In real-world settings, parts of the data may already be (linearly) ordered because of additional background information.
Such data often appears in practice, as especially data containing numerical values (including time) are already ordered.
An ordinal factor analysis should respect such \emph{background orders} as it is unnatural for an analyst to have concepts such as time divided and distributed over different factors.
One could follow (at least) two possible routes to avoid this problem.
The first one is to disregard the ordered part of the data from for the computation of an ordinal factorization.
Then in a second step this order would be added back as an additional factor.
While this guarantees the purity of this additional factor, it might not depict relationships to a sufficient degree.
Thus, one might want to follow a more ambitious path by developing algorithms which do not split such an ordinal structure into different factors but are completely contained in a single factor.
This factor is however still allowed to contain additional information and is contrary to the first approach not restricted to the background order.

Another typical requirement in practice is to update the factorization when data is evolving.
Computing an ordinal factorization of a large dataset will not be (computationally) cheap because of the high computational complexity.
Thus, if a small detail changes in the data it should be possible to integrate it in an existing ordinal factorization without having to repeat the whole computational effort.
Furthermore, the structure of the factorization should be stable to a sufficient degree, as a human data analyst looking a second time at slightly modified data should be presented with familiar ordinal factors.
This proposes two research questions, the first being if one already has to follow additional restrictions when computing the first factorization to allow a later change in the data.
The second is how one can insert the changes into an existing factorization.

\subsubsection{Non-Linear Factor Analysis}

As often complex data is not correlated linearly, further research on non-linear factorizations seems to be promising.
The definition of ordinal factors stemming from a linear order of Boolean factors allows the definition to be extended to factorization into arbitrarily ordered sets.
However, these non-linear factors have to be further restricted as otherwise every lattice is its own, trivial factorization.
We propose to investigate several restrictions for non-ordinal factors that seem reasonable, such as planarity, two-dimensionality, distributivity, and trees.
Another possible approach to gain a reasonable factorization is to fix the number of factors beforehand and to allow arbitrary non-ordinal factors that should approximately be of similar cardinality.
For all those restrictions no preliminary work is available, making this endeavor highly challenging because of anticipated potential complexity as well as visualization problems.

\section{Visualizing, Exploring and Explaining Ordinal Data}
\label{sec-exploration}
Real-world ordinal data are usually too large to be analyzed by simply browsing the set along the edges of the order relation.
The typical means of presenting ordered sets to humans is via line diagrams. A surprisingly hard conceptual problem that we will discuss in this section is the spe\-ci\-fi\-ca\-tion of HCI-founded, formal
optimization criteria for graph drawing, for which one can then develop \fett{efficient layout algorithms} \cite{Battista.1998,durrschnabel2019drawing}.
A second conceptually demanding problem --- still addressed manually today --- is the automatic break-down of large ordered sets into smaller, visualizable parts together with suitable means for their \fett{interactive exploration} \selfcite{stumme99hierarchies}. We will motivate the development of new interaction paradigms based on the decomposition and factorization methods discussed in Sections \ref{sec-algebraic-constructions} and \ref{sec-ordinal-factor-analysis}. These are particularly challenging as they also have to transmit the information about the effects caused by approximations.

\subsection{State of the Art and Open Questions}

Graph Drawing has a long research history, represented a.\,o.\ by the series of Graph Drawing Symposia\footnote{\url{https://dblp.uni-trier.de/db/conf/gd/}} since 1992.
Partial orders and lattices are specific types of graphs, and their drawings can benefit from cycle freeness. Surprisingly, though, there exist only a few  criteria for formalizing readability (\eg maximizing distances between nodes and lines, maximizing angles of crossing lines, minimizing the number of different edge directions, organizing nodes in layers). There is agreement in the community that further criteria are needed, and that they have to be empirically evaluated from the HCI perspective.

\randbemerkung{Order Diagram}
In order theory, a common tool for visualizing and investigating ordered set is the \fett{order diagram}, that is sometimes also called \emph{line diagram} or \emph{Hasse diagram}.
The order diagram of an ordered set $(X,\leq)$ is a directed graph, where the edge $(a,b)\in X\times X$ exists if and only if $a < b$ and there is no $c \in X$ such that $a < c < b$.
In a drawing of this graph the direction of an edge is not signaled by an arrow but by the y-coordinates, with the dot marking the position of $b$ being above the dot marking the position of $a$.
The relative horizontal positions are thus used as visual variable~\cite{Bertin.1967}.
While readable order diagrams can be drawn from an experienced expert by hand, this is not viable in practice, as it is a time-consuming task.
The automatic generation of well interpretable graph drawings is a \fett{surprisingly hard task}. The problem starts with the fact that the readability criteria described above are partly conflicting and that their relative importance varies in different settings.
In the standard work about graph drawing~\cite[Sect. 3.1]{Battista.1998} several divide-and-conquer algorithms for drawing trees, which are closely related to order diagrams, are described.

\randbemerkung{Drawing orders}
In the case of a planar graph (which can be checked in linear time \cite{Hopcroft.1974}), a drawing only consisting of straight lines without bends or curves always exists~\cite[Sect. 4.2 \& 4.3]{Nishizeki.2004} and should be preferred.
For an order diagram with a unique maximum and minimum it can be checked whether it is planar and in this case an upward planar drawing can be computed in linear time~\cite[Sect.~6]{Battista.1998}.
However, while lattices by definition contain a maximum and minimum element, they are usually not planar if they are derived from real world data \cite{Albano.2017}.
A drawing algorithm~\cite[Sect.~3.2]{Battista.1998} for ``serial parallel graphs'', a special family of planar, acyclic graphs straight-line drawings can be equipped with a modification to produce symmetries based on the automorphism group~\cite{Hong.2000}.
Sugiyama et al.~\cite{Sugiyama.1981} published in 1981 an algorithm framework to compute upwards layered drawings of acyclic graphs.
Force directed algorithms %
were introduced in~\cite{Eades.1984} and further refined in \cite{Fruchterman.1991}.
In~\cite{Freese.2004,DBLP:conf/icfca/DurrschnabelS21} this idea is applied to order diagrams. %
Our work~\selfcite{stumme95geometrical} proposes a geometrical heuristic to support a human to draw order diagrams manually.
By trying to emphasize the intrinsic structure of ordered sets our work~\selfcite{durrschnabel2019drawing} proposes a drawing algorithm based on the computation of two linear extensions.
The algorithm is included as a tool in the software-suite \texttt{conexp-clj}\footnote{\url{https://github.com/tomhanika/conexp-clj}} under the name \texttt{DimDraw}~\selfcite{conf/icfca/DurrschnabelHS19}.

\begin{figure}[t]
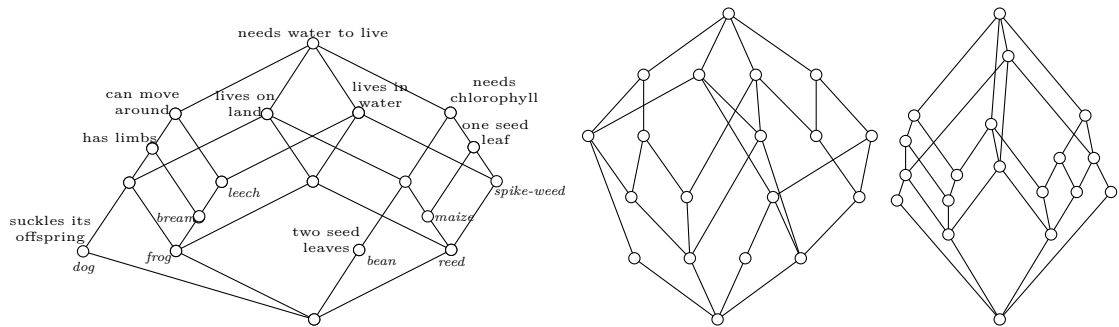

    \centering
    \includegraphics[height=4cm]{tikz/fische_hand.tikz}
    \null\hfill
    \includegraphics[height=12em]{tikz/fische_sugy.tikz}
    \hfill
    \includegraphics[height=12em]{tikz/fische_dimdraw.tikz}
    \hfill\null
    \caption{Three order diagrams for the visualization of the same ordered set: hand-drawn by an expert, by the Sugiyama algorithm and by the DimDraw algorithm.}
    \label{dimdraw}
\end{figure}

\begin{example}

    The example in Figure \ref{dimdraw} shows that it is possible to achieve competitive order diagrams using this approach. The concept lattice was created for structuring an educational film about living beings and water \cite{takacs1984,ganter99formal}. The left diagram was hand-drawn by an FCA expert.
    The diagram in the middle was generated with the algorithm of Sugiyama et al.~\cite{Sugiyama.1981} while the right one was drawn with  the \texttt{DimDraw}-algorithm~\selfcite{durrschnabel2019drawing} which leverages the dimensional structure of the order. Experiments have shown that the DimDraw results are often rather close to hand-drawn diagrams. \qed

\end{example}

As ordered sets and lattices even of modest size (from, say, 30-50 elements upwards) are hard to visualize such that they are still beneficial for a data analyst, further means of interaction are necessary for larger ordered sets. One way is to compute a diagram in 3D, which can be done using~\cite{Freese.2004} or a modification of~\cite{durrschnabel2019drawing}, and then allow to rotate it on the screen, but this does not increase the size of presentable nodes significantly. Another way is to display only a single node together with its upper and lower neighbors at a time, and to shift the focus when the user selects one of these neighbors. This has for instance been used by Carpineto and Romano for Information Retrieval \cite{carpineto04concept}.
A third approach follows the \fett{divide \& conquer} paradigm. Its first realization for concept lattices decomposed the lattice in two or more factors of a sub-direct semi-product, and combined their smaller diagrams in a \fett{nested line diagram} \cite{Wille89d,stumme96local,stumme99hierarchies}. This is also the archetype for the kind of algebraic constructions that we suggested studying in Section~\ref{sec-algebraic-constructions}. This approach became part of a model for \fett{Conceptual Information Systems}
\randbemerkung{Conceptual information systems}%
\selfcite{becker00conceptual} and has been implemented in the \emph{ToscanaJ tool suite}\footnote{\url{http://toscanaj.sourceforge.net/}} \cite{becker02toscana,becker2005toscanaj,dau2011extension}. ToscanaJ also allows, by means of conceptual scaling, to visualize concept lattices derived from data types of any kind, and to navigate intuitively through the data by switching between the conceptual scales and zooming into concepts. However, the initialization of the system requires conceptual and manual work, in particular a meaningful selection of the factors and a manual layout of the order diagram for each factor. The model of Conceptual Information Systems has been generalized in various ways: as ordinal version of \fett{Online-Analytical Processing} \selfcite{stumme98online,stumme00conceptual,eklund00contextual,hereth01reverse,stumme05finite}, for accessing databases and conceptual graphs \cite{KOTTERS2020144} and knowledge bases \cite{Ferre2020}.
A good survey over navigation paradigms and layout algorithms for concept lattices with hybrid data is provided by  \cite{conf/icfca/EklundV10}.

\subsection{Promising Research Questions}
The main objective of the visualization of ordinal structures is to \fett{support human analysts in analyzing and exploring large ordinal data}. In particular, we emphasize on the urgent need of new layout algorithms \fett{that allow for a completely automatic, well interpretable visualization of medium-sized ordered sets and lattices}. These might include \fett{new interaction paradigms based on the decomposition and factorization methods} as discussed in Sections \ref{sec-algebraic-constructions} and \ref{sec-ordinal-factor-analysis}. This approach is particularly challenging as it also has to \fett{transmit the information about which aspects of the data are not represented in the visualization}, due to an approximation.

These (static) visualizations may pave the way for (dynamic) interactive browsing and exploring of complete datasets. Again, the opportunities for new browsing paradigms may benefit from advances in the field of algebraic decompositions and ordinal factor analysis as discussed in Sections \ref{sec-algebraic-constructions} and \ref{sec-ordinal-factor-analysis}.

\subsubsection{Visualization of Ordered Sets and Lattices}
Ordinal factor analysis, as discussed in Section \ref{sec-ordinal-factor-analysis}, will provide a set of chains that cover all points represented in a dataset. Using ideas from~\selfcite{durrschnabel2019drawing}, this could be applied for a novel drawing algorithm, which, however, would not reduce the --- typically too large --- order dimension.
One could hence exploit means for order dimension reduction (towards dimension 2) as discussed in Section \ref{sec-algebraic-constructions}, or employ force directed drawing algorithms in three or more dimensions. The challenge is to incorporate the order constraints such that they are all reflected properly in the resulting diagram.
Another potential approach might be machine-learning-based dimension reduction. This is challenging since the requirement that the $y$-coordinates of data points have to obey the order relation is not encoded in non-linear methods such as t-SNE~\cite{Maaten.2008}, in contrast to linear methods such as PCA~\cite{Wold.1987}. This is also true for deep learning procedures, which nevertheless might be worth being adapted to ordinal data.

\randbemerkung{Partially distributive}
For some special families of lattices (such as distributive lattices, two-dimensional lattices or Boolean lattices), methods exist for generating readable drawings. Thus, in a third line of research one could explore modifications (removing or adding vertices and/or edges) such that the lattice `partly fulfills' these properties --- which first needs to be defined. In a second step, the drawing of this structurally simpler part could then be used to derive a readable drawing for the whole lattice.
A supporting study of this endeavor would be to empirically investigate how large the two-dimensional, distributive or Boolean parts of real-world lattices have to be to justify the applicability of such an algorithm.

\subsubsection{Browsing and Exploring Large Orders}
\randbemerkung{OLAP}
There may be many ways for future approaches to browsing and exploring ordered sets. We assume that some of them will be based on the outcomes of research along the lines described in Sections \ref{sec-algebraic-constructions} and \ref{sec-ordinal-factor-analysis}.  From today's perspective we anticipate these approaches as follows.
The \fett{slice \& dice} paradigm of On-Line Analytical Processing (OLAP) provides a good scheme for analyzing multidimensional numerical data, and first applications to ordinal data exist \selfcite{stumme98online,stumme00conceptual,eklund00contextual,hereth01reverse,stumme05finite}. A similar approach might be followed for the clusterings, decompositions and factorizations developed as described in Sections \ref{sec-algebraic-constructions} and \ref{sec-ordinal-factor-analysis}. To date it is not obvious, though, whether the same visualization and interaction paradigms can also be applied to the new constructions. What is definitely missing in the preliminary work is the handling of disturbances. In a second step, we might therefore address the visualization of and interaction with algebraic clusterings, decompositions and factorizations \fett{that have been approximated}. As there does not exist any preliminary work on this, this part of the research will be of non-incremental nature. Our assumption is that the analyst will be informed about the existence of any non-covered and eventually wrongly assigned data, at least on request. Following the paradigm of \fett{Explainable AI}, the analyst should be enabled to drill down to the original data if necessary, and to obtain explanations in case the construction could not fully respect the data.

\section{Challenges and Long-Term Perspective}
While a wide range of methods for Data Analysis and Knowledge Discovery has been developed for data that are on ratio level (\ie can be adequately modeled with real numbers), there are fewer methods for data on ordinal level. In this paper, we call for a joint activity to bring these methods together and to extend them with new ones, and to establish thus the new research field of Ordinal Data Analysis. The amount of ordinal data in the real world is large, and so is the requirement for adequate methods and algorithms. There are many research opportunities in this field, and we can provide only some teasers in this paper.
We are looking forward to all kinds of contributions to the new field of Ordinal Data Science, both within and outside the five subfields discussed in the previous sections.

We see Ordinal Data Science in a long tradition of Conceptual Knowledge Processing, where we aim at a highly human-centered process of asking, exploring, analyzing, interpreting, and learning about the data in interaction with the underlying database \cite{hereth03conceptual,stumme03off}.  In this line, we follow R. S. Brachman et al. \cite{brachman1993integrated}, who introduced the notion of Data Archaeology for knowledge discovery tasks in which a precise specification of the discovery strategy, the crucial questions, and the basic goals of the task have to be elaborated during such an unpredictable interactive exploration of the data.

We expect that all future contributions to this new field will share a range of challenges that are specific for research on the intersection of computer science and mathematics, which both come with different philosophical foundations, research aims and working paradigms. One challenge will be to develop sound theoretical theories, another one to find computationally feasible solutions and/or approximations. Last but not least the algorithms and methods have to be made accessible to their users, \eg by suitable visualizations, interaction paradigms and human-computer interfaces.

\bigskip
\noindent
We will conclude this paper by illustrating, for the five fields that we discussed in the previous sections, the challenges and long-term perspectives that we expect to face:

\randbemerkung{New flavor of measurement theory}
\medskip
We expect to meet the largest theoretical challenges in the endeavor of developing an ordinal version of the Representational Theory of Measurement. The replacement of the numerical relational structure in RTM by ordinal relational structures (and in the long run by other, non-ordinal relational structures such as symmetry groups) implies the loss of focus on additivity. A generalization of RTM thus requires a much more flexible framework that allows for a large variety of (numerical and) non-numerical scales, and it is not obvious a priori how such a framework might look like.
\randbemerkung{Improved applicability of RTM to psychology}%
A major criticism to RTM is that in practice the pre-conditions for applicability of measures on interval or ratio scale are hard to verify, and that nominal measurements (whose pre-conditions are easier to verify) do not provide enough insight.
When we succeed, we expect a new ordinal theory of measurement to fill the gap between these extremes and to become a significant contribution to the still intense discussion about the applicability of RTM for psychological studies (\cite{doi:10.1111/j.2044-8295.1997.tb02641.x,trendler09measurement,doi:10.1177/0959354318788729}, a good survey is \cite{sep-measurement-science}).

\randbemerkung{`Grand unified theory of measurement'}
Generalized ordinal scales are just a special case of allowing arbitrary algebraic structures as ranges of measurements. In the long run we have the ambition to establish a  \fett{`Grand Unified Theory of Measurement'}, which is likely to be of high interest to (the philosophical foundations of) physics (where traditionally the adaptation of measurement to psychological phenomena has been eyed with some suspicion), as it might allow for describing the study of, \eg the symmetries of elementary particles in the same way as their numerical properties.

\randbemerkung{Combination of two structurally different data types}
\medskip
In Section~\ref{sec-metric-ordered-spaces}, we proposed to develop a joint theory for the interplay of metric functions and ordinal relations, based on the archetype of metric measure spaces. While, in the latter, both metrics and measures operate on the set of real numbers, our proposed approach tries to marry two structurally different domains. We might  thus face the task of developing a new paradigm of interaction which has no precedents that can be followed.

\randbemerkung{Large structures, computationally hard problems, and lack of heuristics}
\medskip
In Sections \ref{sec-algebraic-constructions} to \ref{sec-exploration}, we argued for the exploitation of
algebraic structures for data analysis tasks, which unfortunately go along with large size and computationally hard problems. E.g., the size of concept lattices grows in the worst case exponentially in the size of the data, which in the Information Age are already large by themselves;
and determining the order dimension of an ordered set is \NPhard{}. In Data Science, the typical approach is to use projections, decompositions, approximations, and heuristics. However, these kinds of approximations and heuristics are no subject of study in classical universal algebra, and still have to be established.
\randbemerkung{Algebraic operations and HCI}
In particular,  we suggest to make use of algebraic composition and decomposition techniques for a divide \& conquer approach for browsing and exploring the data. The transformation of the (approximated) algebraic constructions to human--computer interaction paradigms that support users without knowledge of universal algebra and order theory is challenging, as there is little precedence.
Nonetheless, if this works out, it might provide a whole new family of analysis techniques for ordinal data.

\bibliographystyle{plainurl}
\bibliography{tgdkbib}

\end{document}